\documentclass[preprint,12pt]{elsarticle}
\usepackage[T1]{fontenc}
\usepackage[margin=1in]{geometry}
\usepackage{amsmath,amssymb,amsfonts}
\usepackage{graphicx}
\graphicspath{{figures/}{./}}
\usepackage{booktabs}
\usepackage{enumitem}
\usepackage{array}
\usepackage{url}
\newcommand{\fnthresh}{\ensuremath{\mathrm{fn}^{*}}}
\newcommand{\featnorm}{\ensuremath{\mathrm{fn}}}
\newcommand{\tNC}{\ensuremath{T_{\mathrm{NC}}}}

\makeatletter
\def\ps@pprintTitle{%
  \let\@oddhead\@empty \let\@evenhead\@empty
  \let\@oddfoot\@empty \let\@evenfoot\@oddfoot}
\makeatother
\begin{document}
\begin{frontmatter}

\title{Neural Collapse Dynamics: Depth, Activation, Regularisation, and Feature Norm Thresholds}

\author{Anamika Paul Rupa\corref{cor1}}
\ead{anamikal.rupa@howard.edu}
\cortext[cor1]{Corresponding author.}
\affiliation{organization={Department of Electrical Engineering and Computer Science, Howard University},
            city={Washington, DC}, postcode={20059}, country={USA}}

\begin{abstract}
Neural collapse (NC), the convergence of penultimate-layer features to a simplex
equiangular tight frame, is well characterised at equilibrium, but what governs
\emph{when} and \emph{how fast} it emerges has received little systematic study.
We present a controlled study of NC \emph{onset}, showing it is marked by a single measurable quantity: the mean penultimate
feature norm (\featnorm) reaching a critical
threshold \fnthresh\ specific to each (model,~dataset) pair. Three findings support
this. First, \fnthresh\ concentrates tightly within each pair (coefficient of
variation below 8\% at a fixed collapse criterion) and is largely invariant to
depth, width, and weight decay, which set only the \emph{rate} at which \featnorm\
approaches it, whereas activation shifts the threshold itself. Second, a direct
intervention establishes \fnthresh\ as an attractor of the gradient flow, not a
passive correlate: after the feature norm is rescaled over a tenfold range, it
relaxes to the same value. Third, a six-cell (architecture)$\times$(dataset) grid
reveals a non-additive interaction: the architecture effect on \fnthresh\
ranges from $+22.7\%$ to $+458\%$ depending on the dataset (confirmed by a
matched-optimiser control), so \fnthresh\ must be measured per pair, not
extrapolated. A simplified unconstrained-features calculation rationalises the
magnitude of \fnthresh\ and its insensitivity to weight decay. The temporal
ordering is modest: \featnorm\ crosses \fnthresh\ just before collapse but
co-evolves with it, so \fnthresh\ is a stable landmark of the collapsed state, not
an early-warning signal. Together, these results establish the feature-norm
threshold as a reproducible, largely training-invariant characterisation of neural
collapse onset.
\end{abstract}


\end{frontmatter}

\begin{center}
\begin{minipage}{\linewidth}
\centering
\textbf{Graphical abstract}\\[4pt]
\includegraphics[width=0.98\linewidth]{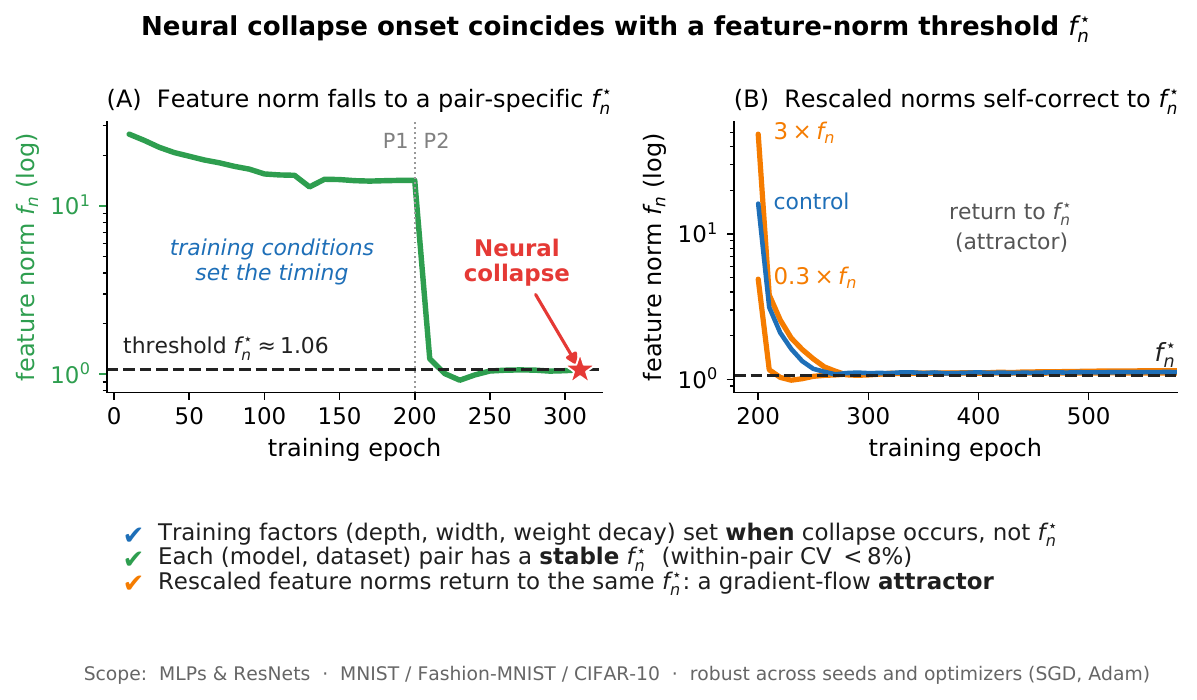}
\end{minipage}
\end{center}

\section{Introduction}
\label{sec:intro}

Neural collapse (NC), first documented by \citet{papyan2020prevalence},
is a striking geometric regularity that emerges at the end of deep network training:
penultimate-layer features converge to a simplex equiangular tight frame~(ETF),
within-class variability vanishes~(NC1), and classifier weights align with class
means~(NC3).
The endpoint of NC is now well characterised theoretically
\citep{zhu2021geometric,han2021neural,zhou2022optimization,jacot2025wide},
yet a fundamental question has received almost no systematic attention:
\emph{what controls when and how fast NC emerges} as a function of architecture
and training hyperparameters?
We answer this empirically: across architectures, datasets, and hyperparameters,
NC onset is consistently marked by the mean feature norm reaching a
model-dataset-specific threshold \fnthresh\ that is largely invariant to training
dynamics, is reached just before collapse, and behaves as an attractor of the
gradient flow, reducing a qualitative dynamics question to a single trackable
quantity.

Understanding collapse timing matters scientifically, by clarifying how
representations reorganise during training, and practically: if the collapsed
state is summarised by a simple, measurable quantity, that quantity could serve as
a compact, inexpensive proxy for collapse geometry across architectures and
optimisation settings.

Our approach is motivated by an analogy with grokking
\citep{power2022grokking,manir2026systematic}, a delayed-generalisation phenomenon
in which \citet{manir2026systematic} showed that timing is governed by the root-mean-square (RMS)
weight norm crossing a model-specific threshold, with weight decay controlling the
approach rate but not the threshold. This motivates an analogous hypothesis: does a
feature-norm threshold play a similar role for collapse onset?

We investigate through five controlled experiments:
\begin{enumerate}[leftmargin=*,noitemsep]
\item Does depth affect NC speed monotonically?
\item Does activation function shape the collapsed geometry, not just the speed?
\item Is there a Goldilocks weight decay zone for NC?
\item Is NC timing predicted by a feature norm threshold, as in grokking?
\item Is the feature-norm threshold \fnthresh\ set by width, architecture, or dataset alone, or by an architecture-dataset interaction?
\end{enumerate}

Our central finding decomposes the dynamics cleanly: training conditions (depth,
activation, weight decay, width) control \emph{when} collapse occurs (\tNC),
whereas \fnthresh\ is tightly concentrated within each (model,~dataset) pair and
\featnorm\ crossing below it consistently precedes collapse onset. Between-pair
differences in \fnthresh\ are substantial but dominated by the ResNet-20/MNIST
cell (ResNet-20 denotes a residual network~\citep{he2016deep}), with the three
multilayer perceptron (MLP-5) cells statistically indistinguishable (Section~\ref{sec:threshold}).

A further notable result comes from the
(architecture)\allowbreak$\,\times\,$\allowbreak(dataset) grid: the effect of
architecture on \fnthresh\ varies substantially with the dataset, so \fnthresh\
does not decompose into independent architecture and dataset contributions
(quantified in Section~\ref{sec:threshold}).

These findings parallel the grokking weight-norm threshold along five structural
axes, with one acknowledged exception (Section~\ref{sec:grokking}); we treat this
as a supporting observation rather than the central claim.

\noindent To our knowledge, this is the first systematic, controlled study of
neural collapse \emph{onset}, as opposed to its already well-characterised
equilibrium geometry. It makes five contributions.
\begin{itemize}[leftmargin=1.4em,itemsep=3pt]
\item \textbf{A new invariant of collapse onset.} We identify a
model-dataset-specific feature-norm threshold \fnthresh: within each
(model,~dataset) pair its value concentrates to a coefficient of variation (CV) below
$8\%$ and is invariant to depth, width, and weight decay, which control only the
\emph{rate} at which \featnorm\ approaches it. Activation is the single exception,
shifting \fnthresh\ itself (Section~\ref{sec:activation}). The concentration
reproduces on five independent seeds in a separate compute environment
($\fnthresh = 1.111 \pm 0.084$, CV~$= 7.6\%$), consistent with the main-grid
MLP-5/MNIST estimate reported in Section~\ref{sec:threshold} ($1.052 \pm 0.066$;
overlapping $95\%$ confidence intervals).
\item \textbf{A consistent temporal ordering at onset.} \featnorm\ reaching \fnthresh\
marks collapse in every confirmed run: \featnorm\ crosses below \fnthresh\ before
the strict NC1 criterion in all $12$ strict-threshold runs (mean lead
$75 \pm 51$~epochs). This ordering is a robust temporal \emph{marker}, not an
early-warning signal. A head-to-head comparison against training loss, accuracy,
weight norm, and gradient norm shows \featnorm\ co-evolves with collapse, so
\fnthresh\ is a stable, pair-specific \emph{landmark} of the collapsed state.
\item \textbf{Interventional, not merely observational, evidence.} Going beyond the
observational and equilibrium-only studies that dominate the NC literature, we
intervene directly: rescaling penultimate features to $0.3\times$ or $3\times$ their
value and resuming training, the feature norm returns to the same \fnthresh\ in
both cases. This establishes \fnthresh\ as an \emph{attractor} of the
mean-squared-error (MSE) plus weight-decay gradient flow, not a passive by-product of collapse
(Section~\ref{sec:intervention}).
\item \textbf{A de-confounded architecture-dataset interaction.} We map the
complete six-cell (architecture)$\times$(dataset) grid and show the architecture
and dataset effects on \fnthresh\ are non-additive (the architecture
effect ranges from $+22.7\%$ to $+458\%$ across datasets). A matched-optimiser
control rules out the obvious confound, confirming the effect is architectural
rather than an artefact of the optimiser pairing (Section~\ref{sec:threshold}).
\item \textbf{A first-principles account of the scale.} A simplified
unconstrained-features analysis derives, in closed form, both the order of
magnitude of \fnthresh\ and its insensitivity to weight decay, giving a
theoretical anchor for the empirical invariant (Section~\ref{sec:ufm}).
\end{itemize}

The remainder of the paper is organised as follows.
Section~\ref{sec:bg} defines NC metrics and the two-phase training protocol.
Section~\ref{sec:related} reviews related work.
Section~\ref{sec:setup} describes the experimental setup.
Section~\ref{sec:results} presents results across five experiments.
Section~\ref{sec:discussion} discusses the grokking parallel, threshold interpretation, and limitations.
Section~\ref{sec:conclusion} concludes.

\section{Background}
\label{sec:bg}

\subsection{NC Metrics}
\label{sec:metrics}

Let $\mathbf{h}_i \in \mathbb{R}^d$ be the penultimate-layer feature for training
example~$i$, $\boldsymbol{\mu}_c$ the class-conditional mean, $\boldsymbol{\mu}_G$
the global mean, and $K$ the number of classes.

\textbf{NC1}: $\text{Tr}(\Sigma_W)/\text{Tr}(\Sigma_B) \to 0$,
where $\Sigma_W$, $\Sigma_B$ are within- and between-class covariance matrices.

\textbf{NC2}: $\frac{1}{K(K-1)}\sum_{i \neq j}|\cos(\mathbf{m}_i,\mathbf{m}_j)
+1/(K{-}1)| \to 0$, where $\mathbf{m}_c = \boldsymbol{\mu}_c - \boldsymbol{\mu}_G$.

\textbf{NC3}: $1 - \frac{1}{K}\sum_c \cos(\hat{\mathbf{w}}_c,\hat{\mathbf{m}}_c)
\to 0$.

NC1 is our primary metric; NC2 and NC3 are reported for the baseline to confirm
all three properties emerge together.
We additionally track mean feature norm
$\featnorm = \frac{1}{N}\sum_i \|\mathbf{h}_i\|_2$
as a dynamical marker of collapse onset.

\textbf{Definition (Feature Norm Threshold).}
We define $\fnthresh = \featnorm$ at epoch $\tNC$, the first epoch with NC1~$< \varepsilon$, 
and ask whether \fnthresh\ is approximately constant across training conditions
within a fixed (model, dataset) pair.
We use ``threshold'' throughout in an operational, descriptive sense:
\fnthresh\ is the \featnorm\ value at which collapse occurs, not a
causal trigger. The intervention experiment (Section~\ref{sec:intervention}) confirms that
\fnthresh\ is a gradient-flow attractor, trajectories above and below
\fnthresh\ both converge to it, so we treat \fnthresh\ as a stable,
pair-specific landmark rather than a mechanism.

\subsection{NC1 Thresholds and Threshold Robustness}
\label{sec:thresholds}

We use NC1~$<0.01$ as the collapse criterion for the positively homogeneous
activations (the rectified linear unit, ReLU~\citep{nair2010relu}, and its leaky
variant, LeakyReLU~\citep{maas2013rectifier})
and NC1~$<0.05$ for the smooth or bounded activations (the hyperbolic
tangent, Tanh~\citep{lecun1998efficient}, and the Gaussian error linear unit,
GELU~\citep{hendrycks2016gelu}) and for shallow networks (depth~$\leq 3$).
The sigmoid linear unit (SiLU)~\citep{elfwing2018sigmoid,ramachandran2017searching}
does not cleanly cross either threshold within the 1000-epoch Phase-2
budget; for SiLU only, we report \featnorm\ at the epoch of minimum Phase-2
NC1 (Section~\ref{sec:activation}). The distinction reflects an
empirical ceiling: the non-homogeneous activations and shallow MLPs saturate
in the NC1 range 0.03--0.08, consistent with weaker global collapse under
these conditions~\citep{kothapalli2022neural}.

A natural concern is whether the resulting differences in \fnthresh\ across
conditions are artefacts of the chosen NC1 levels rather than genuine
collapse-geometry differences. Two arguments rule this out
(Table~\ref{tab:robustness}).

\textbf{Threshold robustness for the strict (NC1~$<0.01$) regime.}
Across six ReLU condition groups (depths 5 and 7, widths 128 and 256, weight
decays $10^{-5}$ and $5\times10^{-5}$), the ratio
\fnthresh(NC1~$<0.02$)/\fnthresh(NC1~$<0.01$) ranges from 0.86 to 1.17 with
mean 1.08. This modest variation is smaller than both the between-pair
gaps (22.7--458\%; Section~\ref{sec:threshold}) and the between-activation gap, so conclusions are insensitive to
the precise NC1 cutoff within this range.

\textbf{Threshold choice for the non-homogeneous regime.}
For GELU, Tanh, and shallow networks, NC1~$<0.05$ is the minimum achievable
threshold given the empirical ceiling, not a free analytical choice. For
SiLU, the minimum Phase-2 NC1 estimator recovers \featnorm\ to within $5\%$
of the mean-over-plateau and end-of-Phase-2 alternatives (Section~\ref{sec:activation}), so
the methodological choice does not affect SiLU's qualitative placement.
The \fnthresh\ differences between activations therefore reflect genuine
equilibrium-geometry differences, not measurement artefacts.

\begin{table}[t]
\centering
\caption{Threshold robustness: \fnthresh\ at NC1~$<0.01$ vs.\ NC1~$<0.02$
(MLP-5, ReLU, MNIST; mean of 3 seeds per condition).}
\label{tab:robustness}
\begin{tabular}{lccc}
\toprule
\textbf{Condition} & \textbf{\fnthresh\ @ NC1$<$0.01} & \textbf{\fnthresh\ @ NC1$<$0.02} & \textbf{Ratio} \\
\midrule
Depth = 5     & 1.077 & 1.251 & 1.16 \\
Depth = 7     & 1.112 & 0.953 & 0.86 \\
Width = 128   & 1.241 & 1.276 & 1.03 \\
Width = 256   & 1.125 & 1.222 & 1.09 \\
$\lambda=10^{-5}$   & 1.035 & 1.197 & 1.16 \\
$\lambda=5\times10^{-5}$ & 0.983 & 1.150 & 1.17 \\
\midrule
Mean ratio    & \multicolumn{3}{c}{$1.08 \pm 0.12$} \\
\bottomrule
\end{tabular}
\end{table}

\subsection{Two-Phase Training Protocol}
\label{sec:twophase}

Following \citet{papyan2020prevalence} and \citet{han2021neural}, we use a
two-phase protocol: \textbf{Phase~1} trains with cross-entropy (CE) loss for
200~epochs to reach $\geq 99\%$ training accuracy; \textbf{Phase~2} switches to
MSE loss with one-hot targets to drive NC1 collapse.
$\tNC$ is the first epoch with NC1 below the applicable threshold
(Section~\ref{sec:thresholds}), reported as an absolute epoch counting from the
start of Phase~1.

We confirm empirically that this protocol is necessary in our setting
(Section~\ref{sec:protocolval}): under CE loss alone, NC1 stalls above 0.076 throughout
600~epochs of training (despite reaching $100\%$ training accuracy from
epoch~10), and \featnorm\ remains an order of magnitude above the equilibrium
value associated with collapse. \citet{han2021neural} establish NC
emergence under CE asymptotically; our result establishes that 600~CE epochs
with Adam and cosine annealing are insufficient in our setup for this
asymptotic regime to be reached.

The protocol's role in our analysis is mechanistic, not foundational. The
threshold rule investigated in this paper concerns the value of \featnorm\ at
which collapse becomes \emph{attainable}, a property of the equilibrium
geometry, not of any particular optimiser. The two-phase protocol functions as
an efficient \emph{mechanism for reaching that equilibrium} within a tractable
compute budget; the threshold rule itself is a statement about where the
equilibrium lies, not about how the network arrives there. The intervention
experiment (Section~\ref{sec:intervention}) reinforces this distinction: \fnthresh\ acts as a
gradient-flow attractor regardless of the trajectory taken to approach it.

\section{Related Work}
\label{sec:related}

\subsection{Neural Collapse: Equilibrium Theory}

Neural collapse was first characterised by \citet{papyan2020prevalence}: in the
terminal phase of training, class means form a simplex equiangular tight frame
(ETF) and within-class variability collapses to zero. The unconstrained-features
model (UFM)~\citep{mixon2020neural,yaras2022neural,zhu2021geometric} treats
penultimate features as free variables and establishes NC as the unique global
minimum under CE and MSE losses with weight decay, with layer-peeled extensions to
deeper models~\citep{ji2022unconstrained}. Further results prove a benign MSE loss
landscape~\citep{zhou2022optimization}, the ETF as the CE global minimum on the
hypersphere~\citep{lu2022neural}, end-to-end results for wide
networks~\citep{jacot2025wide}, NC optimality in ResNets and
Transformers~\citep{sukenik2026neural}, and a mean-field link between NC1 emergence
and loss-landscape geometry~\citep{wu2025neural}. A broad family of losses shares
the same NC optimum~\citep{zhou2022losses}, so the collapsed geometry, and hence the
equilibrium feature norm we study, is largely loss-agnostic, supporting our use of
an MSE phase to reach it efficiently (Section~\ref{sec:twophase}). NC also has
practical utility, from surveys of its principles~\citep{kothapalli2022neural} to
feature collapse predicting transfer accuracy~\citep{li2024transfer}. All of this,
however, concerns the collapsed equilibrium as a fixed point; the phase transition
that triggers onset has remained poorly understood.

\subsection{NC Dynamics}

The dynamics leading to NC have received less attention than the equilibrium.
\citet{han2021neural} give a central-path framework for MSE-driven collapse, and
\citet{wang2025progressive} observe progressive layer-wise collapse in ResNets,
directly relevant to our architecture comparison. Closest to our Goldilocks-zone
finding, \citet{pan2023towards} prove NC emergence under last-layer batch
normalisation and weight decay, and \citet{rangamani2022neural} show, via gradient
flow on the regularised square loss, that weight decay is necessary for collapse,
consistent with our finding that too little or too much prevents it;
\citet{tirer2023perturbation} quantify how regularisation controls closeness to the
collapsed solution. \citet{zhao2026optimizer} show the optimiser, whether stochastic gradient descent (SGD), Adam, or AdamW,
qualitatively shapes NC emergence, proving decoupled weight decay under AdamW
prevents convergence, complementing our finding that excessive coupled weight decay
($\lambda \geq 5\times 10^{-4}$ with Adam) similarly prevents collapse. Multi-layer
UFM extensions~\citep{tirer2022extended,sukenik2023deep} and low-rank bias under
MSE~\citep{garrod2026persistence}, NC under class
imbalance~\citep{gao2023study,hong2024neural,fang2021exploring}, and ETF-based
losses for imbalanced learning~\citep{xie2023neural,yang2022inducing} round out this
line. None of these ask what controls the \emph{timing} of NC across architecture
and hyperparameter choices, or whether a norm threshold marks onset; we address
both.

\subsection{Training Dynamics, Implicit Bias, and Norm Dynamics}

A large body of work studies implicit biases induced by gradient-based
optimisation~\citep{neyshabur2017implicit,soudry2018implicit}.
Early in training, networks often follow trajectories described by the Neural
Tangent Kernel, where features remain close to their initialisation~\citep{jacot2018neural,jacot2025wide}.
As training progresses into the feature-learning regime, representations undergo
significant reorganisation.
Implicit regularisation has been associated with norm minimisation in the
weights~\citep{neyshabur2017implicit,soudry2018implicit}; whether an analogous
norm quantity governs the emergence of representational structure such as neural
collapse is, to our knowledge, open, and motivates the present study.

The closest work on norm-threshold dynamics is the prior study of
\citet{manir2026systematic},
which shows that the RMS weight norm predicts grokking onset in a structurally similar way.
\citet{nanda2023progress} provide mechanistic evidence for progress
measures in grokking; our \featnorm\ threshold is an analogous progress measure for NC.
\citet{liu2022omnigrok} extend grokking to diverse data modalities,
showing that weight norm dynamics are not dataset-specific, a finding we mirror
for NC feature norms.

\subsection{Effects of Architecture and Hyperparameters on NC}

The impact of depth, width, and activation functions on optimisation has been
widely studied, but mostly through loss or accuracy rather than through the
geometry of the learned representation.
Weight decay~\citep{krogh1992simple} is critical in shaping training trajectories, but its interaction
with the geometric emergence of NC, in particular, whether there is a Goldilocks
regularisation regime, has not been systematically characterised.
Our work fills this gap.

\subsection{Contribution Relative to Prior Work}

Table~\ref{tab:positioning} positions this work against representative prior studies
of neural collapse along four axes that define our contribution: whether the work
studies collapse \emph{onset/timing} (not just the equilibrium), proposes a
measurable \emph{norm marker} of onset, varies architecture and dataset
\emph{jointly}, and uses a causal \emph{intervention}.

\begin{table}[t]
\centering
\caption{Positioning relative to representative prior work on neural collapse;
the four comparison axes are defined in the text.}
\label{tab:positioning}
\small
\setlength{\tabcolsep}{3.5pt}
\begin{tabular}{lccccl}
\toprule
Work & Onset & Norm marker & Arch.$\times$data & Interv. & Type \\
\midrule
\citet{papyan2020prevalence}   & No      & No  & No  & No  & Empirical \\
\citet{zhu2021geometric}       & No      & No  & No  & No  & Theory \\
\citet{han2021neural}          & Partial & No  & No  & No  & Theory \\
\citet{pan2023towards}         & Partial & No  & No  & No  & Theory \\
\citet{wang2025progressive}    & Yes     & No  & No  & No  & Empirical \\
\citet{zhao2026optimizer}      & Yes     & No  & No  & No  & Theory+Emp. \\
\textbf{This work}             & \textbf{Yes} & \textbf{Yes} & \textbf{Yes} & \textbf{Yes} & Empirical \\
\bottomrule
\end{tabular}
\end{table}

In contrast, we provide a systematic, controlled study of
NC emergence dynamics across depth, activation, weight decay, and width within a
fixed protocol. Where prior work relates feature- and classifier-norm
\emph{magnitudes} to the collapsed state~\citep{pan2023towards}, we identify a
model-dataset-specific feature-norm \emph{threshold} that is stable within each pair
(CV~$<8\%$ at a fixed activation) and precedes onset in all 12 strict-threshold
runs, holds across the six-cell (architecture~$\times$~dataset) grid and five
activations, and reveals a conditional (non-additive) interaction dominated
by the ResNet-20/MNIST cell. A three-regime weight-decay phase diagram and a
five-way parallel with grokking together suggest norm-threshold dynamics may
underlie delayed representational reorganisation more broadly.

Two recent ICLR~2026 works are closely related but address different
questions. \citet{sakamoto2025explaining} link NC to
grokking in the opposite direction from ours: they use NC1 contraction
to bound grokking generalisation, while we use a grokking-style feature
norm threshold to mark NC onset.
\citet{zhao2026optimizer} study optimiser-dependent NC emergence,
proving that decoupled weight decay under AdamW prevents NC convergence;
our protocol uses coupled weight decay (Adam for MLP, SGD for ResNet) and
documents a separate Goldilocks zone within these optimiser choices. Our
contributions are therefore complementary rather than overlapping with
either line of work.

\section{Experimental Setup}
\label{sec:setup}

\textbf{Datasets.}
MNIST~\citep{lecun1998gradient} (60k/10k, 10~classes),
CIFAR-10~\citep{krizhevsky2009learning} (50k/10k, 10~classes), and
Fashion-MNIST~\citep{xiao2017fashion} (60k/10k, 10~classes; added in revision
as a third dataset to test generalisation of the threshold rule).
No augmentation is applied to CIFAR-10: augmentation inflates within-class
variability and suppresses NC1~\citep{kothapalli2022neural}.

\textbf{Architectures.}
A configurable MLP (width 512 unless stated) and ResNet-20~\citep{he2016deep},
each evaluated on MNIST, Fashion-MNIST, and CIFAR-10 to complete the
architecture~$\times$~dataset grid (Section~\ref{sec:threshold}).
Both use ReLU unless stated.
The MLP body is a \texttt{Flatten} layer followed by $d$ repetitions of a
(\texttt{Linear}, activation) block and a final \texttt{Linear} classification head,
where $d$ is the number of hidden layers (depth).
Weights are initialised with Kaiming normal~\citep{he2015delving};
biases zero.

\textbf{Optimisers.}
Optimiser is tied to architecture in the main grid (a coupling we de-confound
with matched-optimiser runs in Section~\ref{sec:threshold}).
MLP-5 runs use Adam~\citep{kingma2014adam} (lr~$= 10^{-3}$, cosine
annealing~\citep{loshchilov2016sgdr}, default $\lambda = 10^{-4}$).
ResNet-20 runs use SGD with Nesterov momentum~\citep{sutskever2013importance}
(lr~$= 0.1$, momentum~0.9, default $\lambda = 10^{-3}$, with MultiStep LR decay
at Phase-2 epochs 300/450).

\textbf{Hardware.}
All experiments ran on single NVIDIA T4 GPUs via Google Colab and Kaggle.

\textbf{Statistical analysis.}
Each condition uses 3 random seeds (seeds 0--2; 0--4 for GELU, to mitigate
initialisation sensitivity); the intervention experiment (Section~\ref{sec:intervention}) uses
3 seeds per arm, with all seeds fixed.
Point estimates are reported as mean~$\pm$~sample standard deviation, with
coefficients of variation (CV) and all dispersion statistics computed using the
sample standard deviation (ddof~$= 1$). For \fnthresh\ estimates we report 95\%
Student-$t$ confidence intervals (CI; df~$= N-1$); with only 3--4 seeds per cell
we prefer these to bootstrap intervals, which are unreliable at this sample size.
Between-pair differences in \fnthresh\ are assessed with the rank-based
Kruskal-Wallis test~\citep{kruskal1952use} in place of one-way analysis of variance (ANOVA), whose equal-variance assumption
is violated by the near-zero within-pair variance of the ResNet-20 cells
(Section~\ref{sec:threshold}); we report $\varepsilon^2$ as the rank-based effect size.
Trend analyses across the small number of design points (e.g., feature norm or
collapse epoch versus width) are treated as exploratory rather than as formal
tests. NC metrics are computed on the full training set at every epoch.

\section{Results}
\label{sec:results}

We first establish baseline collapse dynamics under the two-phase protocol, then
test the five hypotheses of Section~\ref{sec:intro} in turn: depth (H1),
activation (H2), weight decay (H3), the feature-norm threshold across the
architecture--dataset grid (H4), and width (H5). We then probe the threshold's
causal status with a direct intervention, check that it extends to a Vision
Transformer, and confirm the two-phase protocol is necessary. Throughout we track two
quantities, the epoch of collapse \tNC\ and the feature norm at collapse
\fnthresh. Our central finding is their separation: training factors move \tNC,
whereas \fnthresh\ stays specific to each (model,~dataset) pair.

\subsection{Baseline: Feature Norm Compresses 25-Fold at Collapse}
\label{sec:baseline}

\begin{figure}[htbp]
  \centering
  \includegraphics[width=\linewidth]{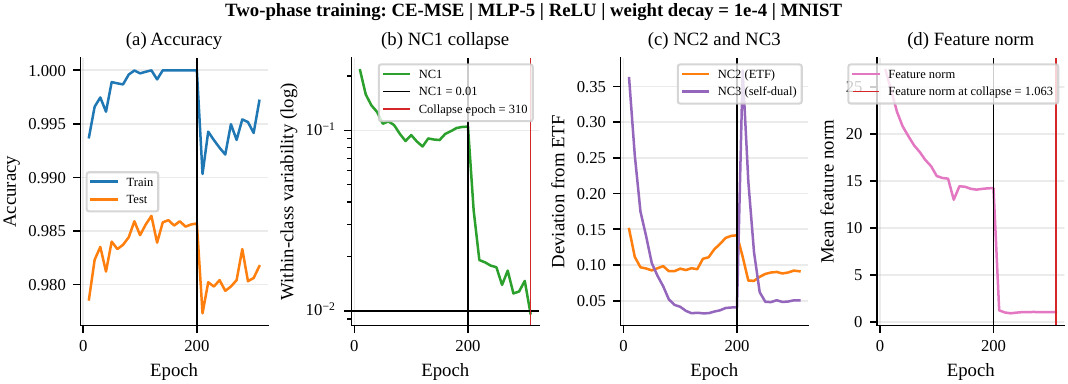}
  \caption{MNIST baseline dynamics (MLP-5, ReLU, $\lambda=10^{-4}$):
  (a)~accuracy, (b)~NC1, (c)~NC2 and NC3, (d)~feature norm.}
  \label{fig:baseline}
\end{figure}

Fig.~\ref{fig:baseline} shows the MNIST baseline (MLP-5, ReLU, $\lambda=10^{-4}$).
NC1 decreases gradually from ${\approx}0.3$ at epoch 10 to ${\approx}0.1$ by
the end of Phase 1 but never approaches the collapse threshold, despite
$100\%$ training accuracy from epoch 10 onward, confirming that the
terminal phase is entered but collapse does not occur under CE alone.
In Phase~2, NC1 collapses to 0.0097 at $\tNC = 310$~epochs.
The feature norm undergoes a $25\times$ compression across the full two-phase
training: from ${\approx}27$ at epoch~10 of Phase~1 to $\featnorm = 1.063$ at $\tNC$.
NC2 and NC3 decline in parallel, confirming all three NC properties emerge together.

\begin{figure}[htbp]
  \centering
  \includegraphics[width=\linewidth]{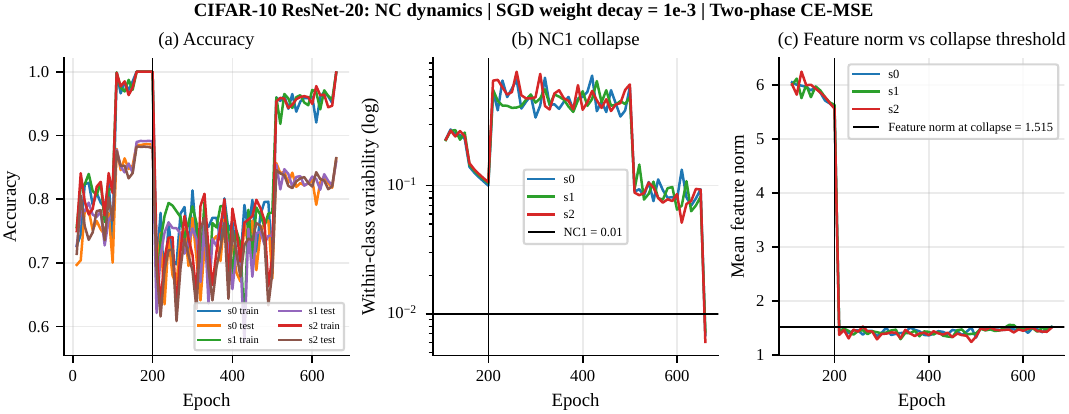}
  \caption{CIFAR-10 ResNet-20 dynamics (3~seeds, SGD, $\lambda=10^{-3}$):
  (a)~accuracy, (b)~NC1, (c)~feature norm per seed.}
  \label{fig:cifar}
\end{figure}

ResNet-20/CIFAR-10 (Fig.~\ref{fig:cifar}) collapses at $\tNC=660$ across all
3~seeds at $\featnorm = 1.515 \pm 0.008$ (CV~$= 0.5\%$).
The \featnorm\ at collapse is 44\% higher than the pooled MLP-5/MNIST value
($1.052$; Section~\ref{sec:threshold}), a gap we decompose in
Experiments~4 and~5 below.

\subsection{H1, Depth: Non-Monotonic Collapse Speed}
\label{sec:depth}

Table~\ref{tab:depth} reports MLP results at depth $\in \{2,3,5,7\}$
(ReLU, $\lambda=10^{-4}$, 3~seeds). Following Section~\ref{sec:thresholds}, the
NC1 collapse criterion is the strict $0.01$ for depth~$\geq5$ and the relaxed
$0.05$ for depth~$\leq3$.
The depth--$\tNC$ relationship is \emph{non-monotonic}: depth-3 collapses fastest
(250~epochs), depth-7 slowest (340~epochs), with depth-2 and depth-5 intermediate.
Greater capacity does not accelerate collapse; intermediate depth appears optimal.
This mirrors the non-monotonic depth finding in grokking~\citep{manir2026systematic}.
One interpretation is that deeper networks accumulate stronger implicit
weight-decay regularisation across more layers, slowing the \featnorm\ trajectory toward
\fnthresh; we treat this as a hypothesis requiring theoretical investigation.
The invariance of \fnthresh\ to depth should be read as holding \emph{at a fixed
collapse criterion}: depths~5 and~7, which both reach the strict NC1~$<0.01$, differ by only
$3\%$ (Table~\ref{tab:depth}), whereas the shallow depths~2 and~3, measured only
at the relaxed NC1~$<0.05$, collapse at higher \featnorm. The larger spread of the shallow rows
therefore reflects the relaxed criterion rather than a genuine depth dependence of
the threshold.

\begin{table}[t]
\centering
\caption{Depth sweep (MLP, ReLU, $\lambda=10^{-4}$, MNIST; mean~$\pm$~std, $N=3$).}
\label{tab:depth}
\begin{tabular}{cclcc}
\toprule
\textbf{Depth} & \textbf{N} & \textbf{NC1 thresh.}
  & $T_{\mathrm{NC}}$ & \textbf{\featnorm\ at $\tNC$} \\
\midrule
2 & 3 & $<0.05$ & $273 \pm 67$  & $1.428 \pm 0.323$ \\
3 & 3 & $<0.05$ & $250 \pm 35$  & $1.417 \pm 0.276$ \\
5 & 3 & $<0.01$ & $297 \pm 12$   & $1.077 \pm 0.025$ \\
7 & 3 & $<0.01$ & $340 \pm 10$  & $1.112 \pm 0.040$ \\
\bottomrule
\end{tabular}
\end{table}

\subsection{H2, Activation: Joint Control of Speed and Collapsed Geometry}
\label{sec:activation}

Table~\ref{tab:act} compares five activations at depth~5
(ReLU, LeakyReLU(0.01), Tanh, GELU, SiLU; $\lambda=10^{-4}$, 3--5 seeds
each; NC1 thresholds per Section~\ref{sec:thresholds}). LeakyReLU and SiLU were
added in revision (3~seeds each); GELU uses five seeds (0--4), all of which
collapse and are retained.

\begin{table}[t]
\centering
\caption{Activation sweep (MLP-5, $\lambda=10^{-4}$, MNIST; mean~$\pm$~std).
$^{\dagger}$SiLU plateaus above NC1~$=0.05$; \featnorm\ is reported at the epoch
of minimum Phase-2 NC1 (Section~\ref{sec:activation}).}
\label{tab:act}
\begin{tabular}{cclcc}
\toprule
\textbf{Act.} & \textbf{N} & \textbf{NC1 thresh.}
  & $T_{\mathrm{NC}}$ & \textbf{\featnorm\ at $\tNC$} \\
\midrule
ReLU       & 3 & $<0.01$                & $297 \pm 12$    & $1.077 \pm 0.025$ \\
LeakyReLU  & 3 & $<0.01$                & $280 \pm 26$   & $1.026 \pm 0.029$ \\
Tanh       & 3 & $<0.05$                & $220 \pm 0$    & $1.325 \pm 0.040$ \\
GELU       & 5 & $<0.05$                & $260 \pm 23$   & $2.118 \pm 0.446$ \\
SiLU       & 3 & plateau$^{\dagger}$    & $1027 \pm 179$ & $3.257 \pm 0.188$ \\
\bottomrule
\end{tabular}
\end{table}

Three patterns stand out.
First, activation jointly shapes both collapse speed and \featnorm\ at $\tNC$.
Across the five activations tested, \featnorm\ at $\tNC$ spans a roughly threefold range
(1.026--3.257). Collapse speed varies more modestly: $\tNC$ ranges from 220 to
297~epochs (a factor of 1.35) across the four activations that cleanly cross a
fixed NC1 threshold, with SiLU's plateau endpoint at 1027~epochs separated by
methodology rather than dynamics.
This contrasts with grokking, where activation changes rate but not the weight
norm threshold~\citep{manir2026systematic}; for NC, each activation appears to
define a different equilibrium feature geometry.

Second, the five activations cluster into two groups by \fnthresh\ magnitude that
align with their positive-homogeneity property. \emph{Positively homogeneous}
activations (ReLU, LeakyReLU(0.01); $f(\alpha x) = \alpha f(x)$ for $\alpha>0$)
collapse at $\featnorm \approx 1.0$ with within-group spread of only $5\%$
and overlapping $95\%$ $t$-CIs (Table~\ref{tab:act}).
\emph{Non-homogeneous} activations span a wider range (Table~\ref{tab:act}): the
bounded Tanh sits above the homogeneous cluster, and the smooth GELU and SiLU
higher still. We discuss possible mechanisms in Section~\ref{sec:mechanisms} (H-M1) and treat the
homogeneity ordering as descriptive only, not causal. Two caveats limit it.
First, isolating the mechanism requires a control that preserves piecewise-linear
shape but removes positive homogeneity, for example a fixed-threshold ReLU,
$\mathrm{ReLU}(x-\tau)$ with $\tau>0$ held constant, which introduces an intrinsic
scale $\tau$ (note that the naive rescaling $\alpha\,\mathrm{ReLU}(x/\alpha)=
\mathrm{ReLU}(x)$ is an identity for ReLU and cannot serve as such a control). Second, the ordering
above compares \fnthresh\ measured at \emph{different} NC1 criteria
(NC1~$<0.01$ for the homogeneous activations, which reach deeper collapse,
versus NC1~$<0.05$ or the plateau estimator for the others). When the
piecewise-linear and bounded activations are instead compared at a single common
criterion (NC1~$<0.05$, which ReLU, LeakyReLU, Tanh, and GELU all reach), the
separation by homogeneity does not persist: ReLU ($1.81\pm0.59$) and
LeakyReLU($0.01$) ($1.68\pm0.26$) overlap with GELU ($2.12\pm0.45$), and the
bounded non-homogeneous Tanh is the \emph{lowest} ($1.33\pm0.04$); SiLU does not
reliably reach NC1~$<0.05$ and plateaus higher still ($\featnorm\approx 3.26$).
The apparent clustering at
\featnorm~$\approx 1.0$ for ReLU/LeakyReLU therefore partly reflects their
reaching the stricter criterion, not positive homogeneity alone. We conclude
that activation strongly affects \fnthresh\ at collapse, spanning a roughly
$1.6\times$ range across the four activations that reach the common NC1~$<0.05$
criterion (and wider still once SiLU's higher plateau is included), but the
specific positive-homogeneity account is not supported once collapse depth is
held constant.

\begin{figure}[htbp]
  \centering
  \includegraphics[width=\linewidth]{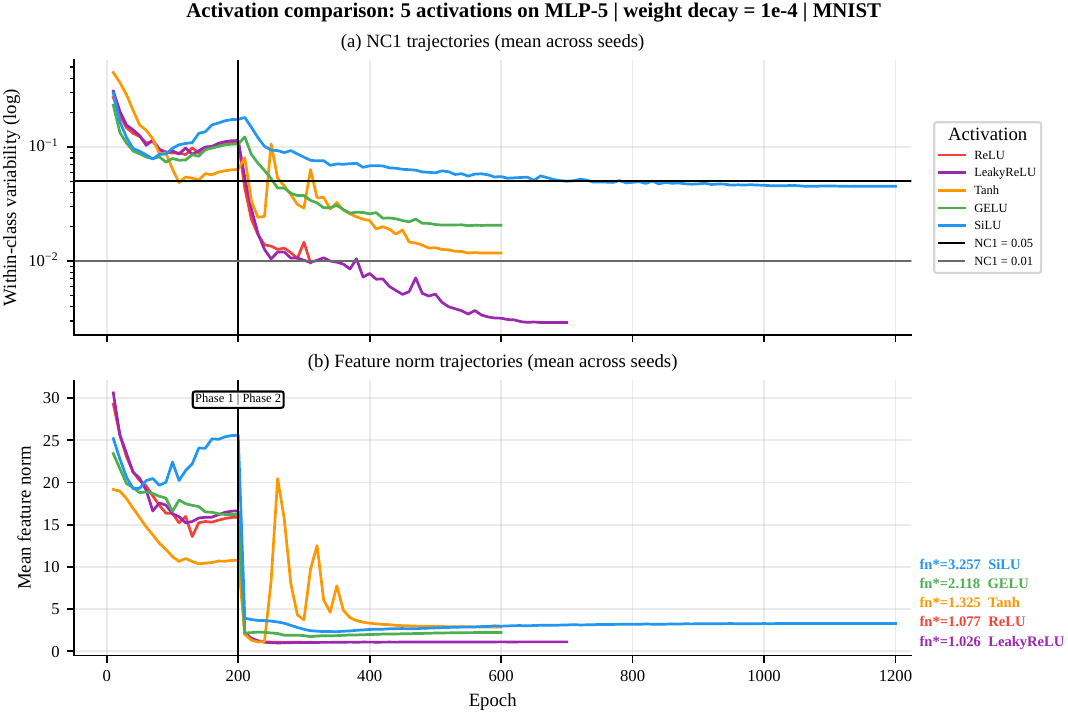}
  \caption{Activation comparison on MLP-5/MNIST ($\lambda=10^{-4}$, 3~seeds):
  (a)~NC1 (log scale), (b)~feature norm (log scale), with per-activation
  \fnthresh\ labelled at the right edge. Tanh NC1 is lightly smoothed
  (3-point rolling mean) for display.}
  \label{fig:act_comparison}
\end{figure}

\begin{figure}[htbp]
  \centering
  \includegraphics[width=\linewidth]{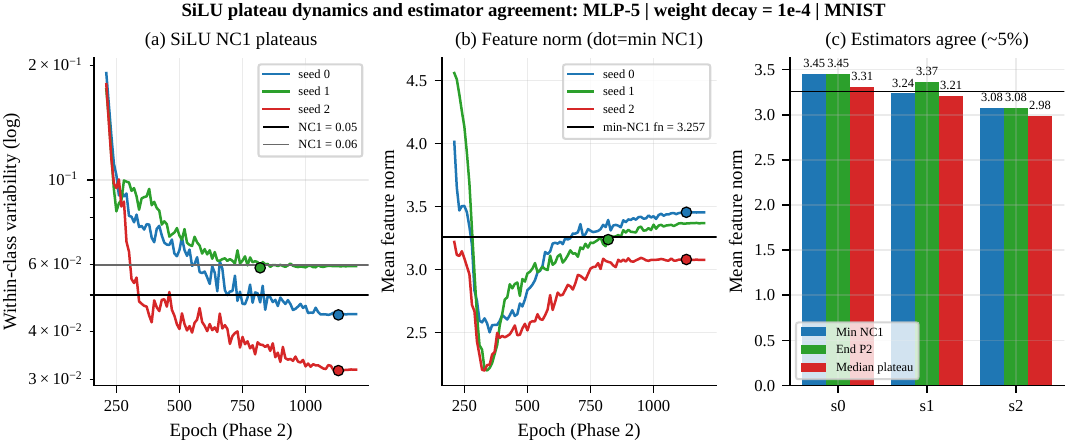}
  \caption{SiLU plateau dynamics (MLP-5/MNIST, $\lambda=10^{-4}$, Phase-2
  $1000$~epochs, 3~seeds): (a)~NC1 plateau, (b)~feature norm with the reported
  \fnthresh\ (dashed), (c)~agreement of three plateau estimators.}
  \label{fig:silu_plateau}
\end{figure}

Third, smooth non-homogeneous activations exhibit higher seed-to-seed variability
and slower / plateauing collapse (Fig.~\ref{fig:act_comparison}).
ReLU, LeakyReLU, and Tanh all have seed-to-seed CV below $3.1\%$ in \featnorm\ at
$\tNC$. GELU's CV is $21\%$, and SiLU's NC1 plateaus just above $0.05$ for all three
seeds rather than crossing a fixed threshold cleanly (Fig.~\ref{fig:silu_plateau}). We hypothesise this
reflects the non-piecewise-linear gradient structure of GELU and SiLU creating a
more complex loss landscape with initialisation-sensitive trajectories.

\subsection{H3, Weight Decay: Rate Control Without Threshold Shift}
\label{sec:wd}

\begin{figure}[htbp]
  \centering
  \includegraphics[width=\linewidth]{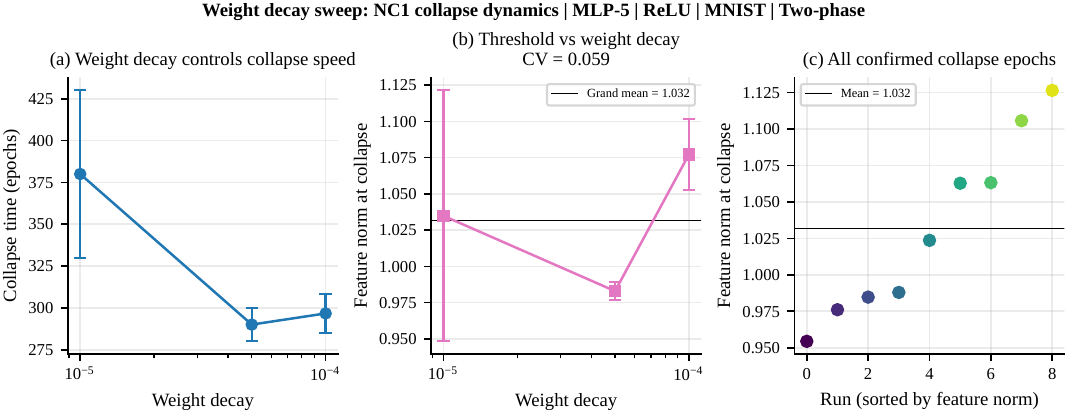}
  \caption{Weight-decay sweep (MLP-5, ReLU, MNIST):
  (a)~$\tNC$ vs.\ $\lambda$, (b)~\featnorm\ at $\tNC$, (c)~all confirmed values.}
  \label{fig:wd}
\end{figure}

Table~\ref{tab:wd} and Fig.~\ref{fig:wd} report results for
$\lambda \in \{10^{-5}, 5\times10^{-5}, 10^{-4}, 5\times10^{-4}\}$
(MLP-5, ReLU, 3~seeds, NC1~$< 0.01$).

\begin{table}[t]
\centering
\caption{Weight-decay sweep (MLP-5, ReLU, MNIST, NC1~$<0.01$; mean~$\pm$~std).
DNF (did not finish): NC1 stalled at ${\approx}0.03$ for all seeds.}
\label{tab:wd}
\begin{tabular}{lccrc}
\toprule
$\lambda$ & \textbf{N} & \textbf{Collapsed}
  & $T_{\mathrm{NC}}$ & \textbf{\featnorm\ at $\tNC$} \\
\midrule
$10^{-5}$        & 3 & 3/3 & $380 \pm 50$ & $1.035 \pm 0.087$ \\
$5\times10^{-5}$ & 3 & 3/3 & $290 \pm 10$ & $0.983 \pm 0.006$ \\
$10^{-4}$        & 3 & 3/3 & $297 \pm 12$  & $1.077 \pm 0.025$ \\
$5\times10^{-4}$ & 3 & 0/3 & DNF          & n/a              \\
\bottomrule
\end{tabular}
\end{table}

\begin{table}[htbp]
\centering
\caption{Feature-norm threshold \fnthresh\ pooled across all MNIST/MLP-5/ReLU
conditions; the pool mixes strict (NC1~$<0.01$) and relaxed (depths~2--3,
NC1~$<0.05$) collapse criteria.}
\label{tab:allconds}
\small
\setlength{\tabcolsep}{5pt}
\begin{tabular}{llcc}
\toprule
\textbf{Condition} & \textbf{Values varied} & $T_{\mathrm{NC}}$ range & \textbf{\fnthresh\ (mean $\pm$ std)} \\
\midrule
Depth sweep    & depth $\in \{2,3,5,7\}$                         & 250--340  & $1.259 \pm 0.251$ \\
WD sweep       & $\lambda \in \{10^{-5},\ 5\times10^{-5},\ 10^{-4}\}$ & 290--380 & $1.032 \pm 0.061$ \\
Width sweep    & width $\in \{128, 256, 512, 1024\}$             & 257--383  & $1.135 \pm 0.085$ \\
\midrule
All ReLU runs  & (combined, $N=21$)                              & 250--383  & $1.093 \pm 0.092$ \\
\bottomrule
\end{tabular}
\end{table}

\begin{table}[htbp]
\centering
\caption{Feature-norm threshold \fnthresh\ by (model, dataset) pair (ReLU);
mean~$\pm$~std, with the NC1 collapse criterion and confirmed-collapse count $N$.
RN-20: ResNet-20; F-MNIST: Fashion-MNIST.}
\label{tab:threshold}
\footnotesize
\setlength{\tabcolsep}{3pt}
\begin{tabular}{llccccc}
\toprule
\textbf{Dataset} & \textbf{Architecture} & \textbf{NC1}
  & $N$ & \textbf{\fnthresh} & \textbf{95\% $t$-CI}
  & \textbf{CV} \\
\midrule
MNIST         & MLP-5     & $<0.01$ & 12 & $1.052 \pm 0.066$ & $[1.010,\ 1.093]$ & $6.2\%$ \\
MNIST         & ResNet-20 & $<0.01$ &  3 & $5.867 \pm 0.034$ & $[5.781,\ 5.952]$ & $0.6\%$ \\
Fashion-MNIST & MLP-5     & $<0.05$ &  3 & $1.367 \pm 0.077$ & $[1.176,\ 1.559]$ & $5.6\%$ \\
Fashion-MNIST & ResNet-20 & $<0.01$ &  3 & $1.678 \pm 0.012$ & $[1.648,\ 1.708]$ & $0.7\%$ \\
CIFAR-10      & MLP-5     & $<0.05$ &  3 & $0.900 \pm 0.065$ & $[0.738,\ 1.063]$ & $7.3\%$ \\
CIFAR-10      & ResNet-20 & $<0.01$ &  3 & $1.515 \pm 0.008$ & $[1.494,\ 1.535]$ & $0.5\%$ \\
\midrule
\multicolumn{4}{l}{Arch.\ effect, MNIST (MLP-5$\to$RN-20)}
  & \multicolumn{3}{l}{$\Delta = +4.815$ \ $(+458\%)$} \\
\multicolumn{4}{l}{Arch.\ effect, F-MNIST (MLP-5$\to$RN-20)}
  & \multicolumn{3}{l}{$\Delta = +0.311$ \ $(+22.7\%)$} \\
\multicolumn{4}{l}{Arch.\ effect, CIFAR-10 (MLP-5$\to$RN-20)}
  & \multicolumn{3}{l}{$\Delta = +0.614$ \ $(+68\%)$} \\
\multicolumn{4}{l}{Dataset effect, MLP-5 (MNIST$\to$CIFAR-10)}
  & \multicolumn{3}{l}{$\Delta = -0.152$ \ $(-14\%)$} \\
\multicolumn{4}{l}{Dataset effect, RN-20 (MNIST$\to$CIFAR-10)}
  & \multicolumn{3}{l}{$\Delta = -4.352$ \ $(-74\%)$} \\
\multicolumn{7}{l}{\textit{Effects are dataset- and architecture-dependent; they do not add.}} \\
\bottomrule
\end{tabular}
\end{table}

Three findings emerge, together forming a \emph{phase diagram} in $\lambda$:
\textbf{(i)~Too-high $\lambda$ prevents collapse entirely.}
$\lambda = 5\times10^{-4}$ pins the feature norm above \fnthresh\ throughout Phase~2
(NC1 stalls at ${\approx}0.03$), a \emph{frozen} phase in which the network is
too heavily regularised for collapse to occur.
\textbf{(ii)~Collapsing $\lambda$ values show rate control.}
Across the three collapsing values ($\lambda \in \{10^{-5}, 5\times10^{-5}, 10^{-4}\}$),
$\tNC$ ranges from 290 to 380~epochs, a 90-epoch window from a $10\times$ change in $\lambda$.
Lower $\lambda$ slows the \featnorm\ decay trajectory, delaying collapse; higher $\lambda$
(within the collapsing regime) accelerates it.
The optimal $\lambda$ for fastest collapse in our setup is $5\times10^{-5}$
($\tNC = 290$~epochs), not the smallest tested value.
\textbf{(iii)~\featnorm\ at $\tNC$ is stable across $\lambda$:} the grand mean is
$1.032$ (CV~$= 5.9\%$ across 9~confirmed runs; Table~\ref{tab:allconds}).
The three regimes together define a Goldilocks structure: too little regularisation
slows collapse, the optimal range produces fastest collapse, and too much prevents
it entirely, while the collapse-associated \featnorm\ value remains invariant throughout.

\subsection{H4, Feature Norm Threshold: A Stable, Pair-Specific Invariant}
\label{sec:threshold}

\begin{figure}[htbp]
  \centering
  \includegraphics[width=\linewidth]{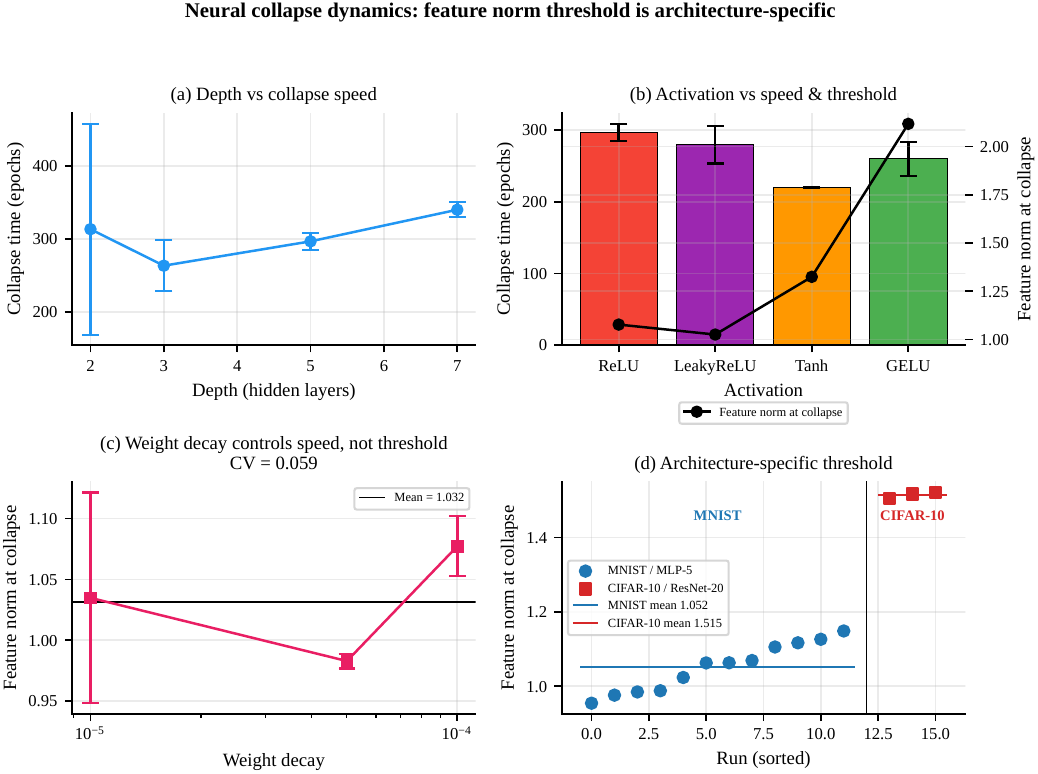}
  \caption{Summary of NC dynamics results: (a)~depth effect on $\tNC$,
  (b)~activation effect, (c)~weight-decay effect, (d)~\featnorm\ by
  (model, dataset) pair.}
  \label{fig:summary}
\end{figure}

Table~\ref{tab:threshold} and Fig.~\ref{fig:summary} consolidate all confirmed
$\tNC$ results under ReLU.

Within each pair, \fnthresh\ concentrates tightly (CV~$<8\%$ in all cases),
reinforcing that the characteristic value is a property of the (model, dataset) pair,
not of the training path approaching it.

Between-pair variation is significant by a Kruskal-Wallis test
($H=15.75$, $p=0.0013$, $\varepsilon^2=0.75$ across the four original pairs;
$H=23.3$, $p<0.001$, $\varepsilon^2=0.87$ across all six cells); we use a
rank-based test because the near-zero
within-group variance of the ResNet-20 cells violates the equal-variance
assumption of one-way ANOVA. This variation is dominated by the
ResNet-20/MNIST cell, the three MLP-5 cells have overlapping 95\% CIs
(Table~\ref{tab:threshold}) and are not individually separable.

\begin{figure}[htbp]
  \centering
  \includegraphics[width=\linewidth]{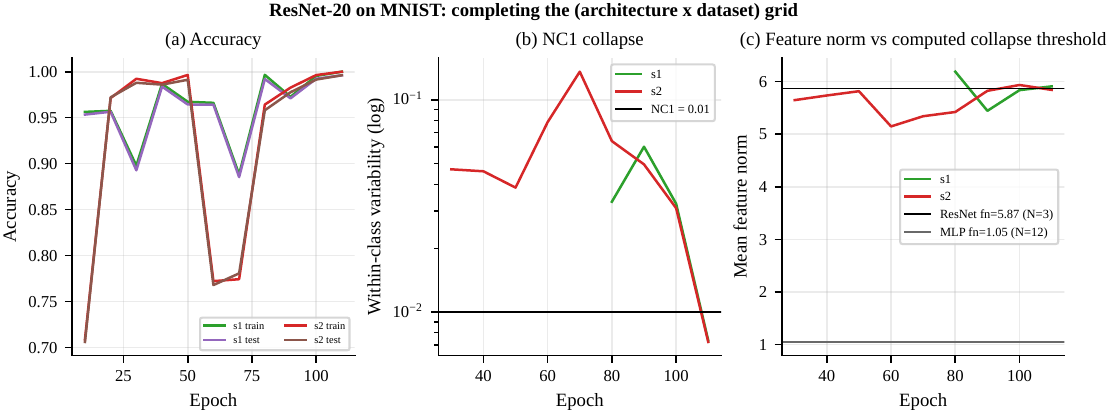}
  \caption{ResNet-20 on MNIST (3~seeds): (a)~accuracy, (b)~NC1,
  (c)~feature norm per seed.}
  \label{fig:resnet_mnist}
\end{figure}

To populate the second architecture row, we run ResNet-20 on MNIST
(Fig.~\ref{fig:resnet_mnist}). All three seeds collapse at $\tNC = 110$
epochs, roughly six times faster than ResNet-20 on CIFAR-10 (660~epochs),
consistent with MNIST being a much simpler task, with $\fnthresh =
5.867 \pm 0.034$ (CV~$=0.6\%$).

The completed grid reveals a non-additive architecture $\times$ dataset interaction
across the six cells evaluated (Table~\ref{tab:threshold}).
The architecture effect (MLP-5$\to$ResNet-20) varies across all three
datasets: $+458\%$ on MNIST, $+22.7\%$ on Fashion-MNIST, and $+68\%$ on
CIFAR-10. The MNIST architecture effect is therefore substantially
larger than the effects on either Fashion-MNIST or CIFAR-10, both of which
sit in the $+20$--$70\%$ range.
The dataset effect (MNIST$\to$CIFAR-10) is $-14\%$ \emph{conditional on
MLP-5} but $-74\%$ \emph{conditional on ResNet-20}, a $5.3\times$
difference depending on the architecture.
These effects do not add: there is no single ``architecture contribution''
or ``dataset contribution'' to \fnthresh.
Even on a log scale, a multiplicative model under-predicts the
ResNet-20/MNIST value by $232\%$, confirming the interaction is not simply
a scaling artefact.
\fnthresh\ is therefore a joint property of the (model, dataset) pair,
not decomposable into independent factors.

\begin{figure}[htbp]
  \centering
  \includegraphics[width=\linewidth]{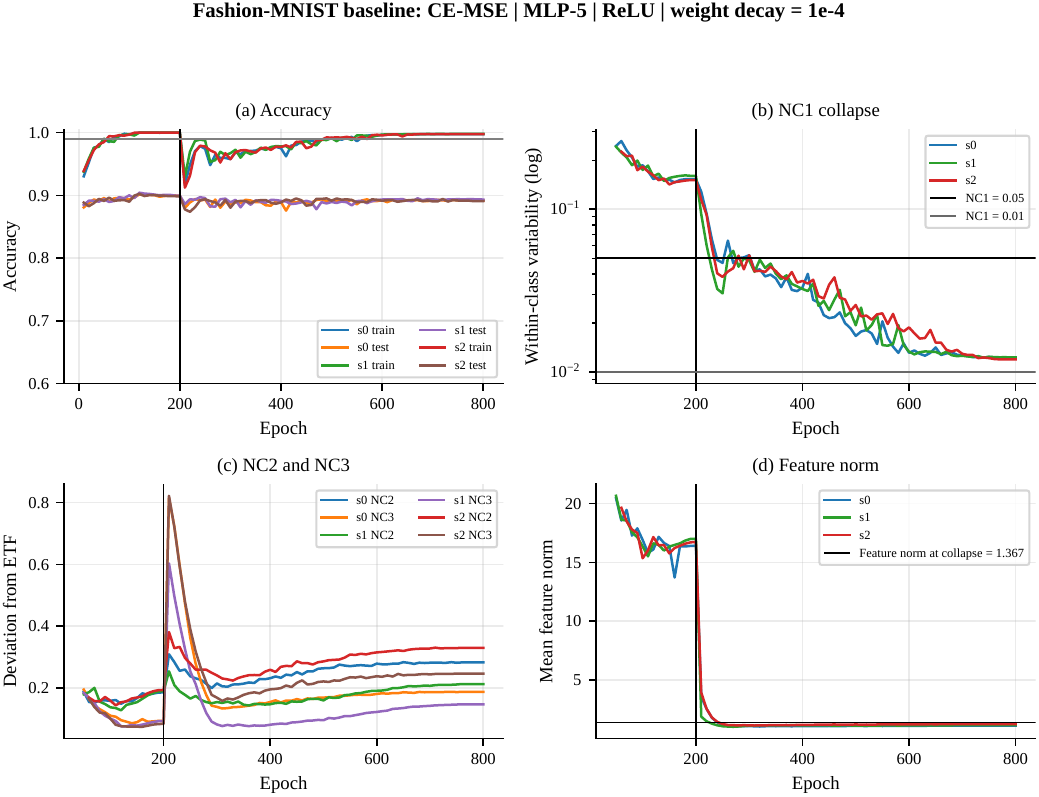}
  \caption{Fashion-MNIST baseline dynamics (MLP-5, ReLU, $\lambda=10^{-4}$;
  mean~$\pm$~std, 3~seeds): (a)~accuracy, (b)~NC1, (c)~NC2 and NC3,
  (d)~feature norm (dashed line marks the reported \fnthresh).}
  \label{fig:fmnist_baseline}
\end{figure}

\begin{figure}[htbp]
  \centering
  \includegraphics[width=\linewidth]{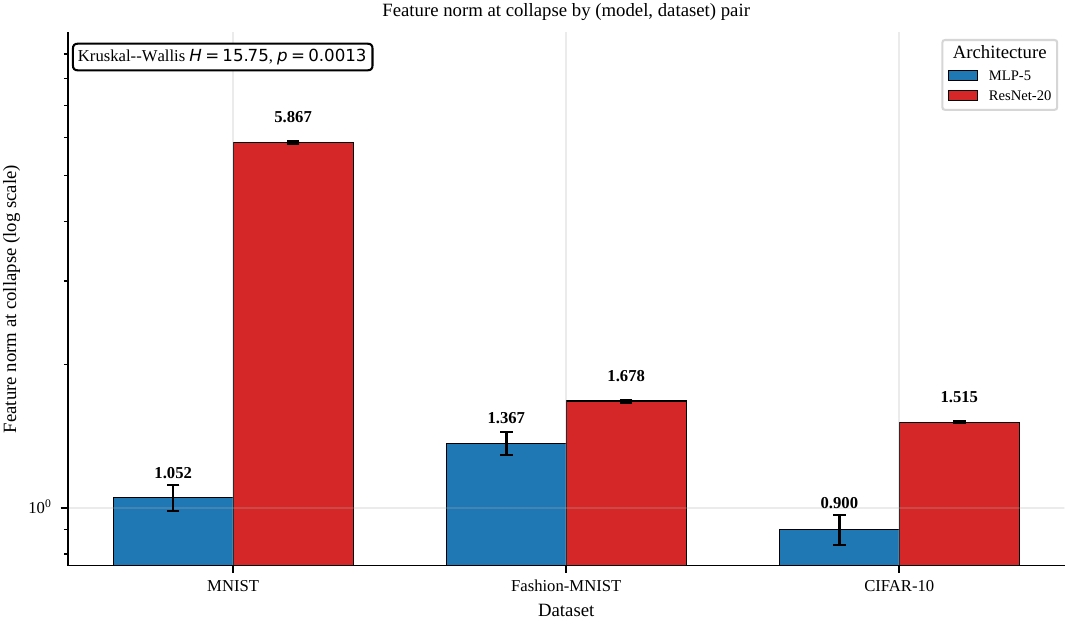}
  \caption{Feature-norm threshold \fnthresh\ by (model, dataset) pair across the
  six evaluated cells, grouped by dataset and coloured by architecture
  (mean~$\pm$~std).}
  \label{fig:summary_panel_d}
\end{figure}

\textbf{Generalisation to a third dataset.}
To address whether the within-pair concentration rule depends on the specific
choice of MNIST and CIFAR-10, we extend both rows of the
architecture~$\times$~dataset grid to a third dataset,
Fashion-MNIST~\citep{xiao2017fashion}. For MLP-5/Fashion-MNIST,
$\featnorm$ at NC1~$<0.05$ is $1.367 \pm 0.077$ (CV~$=5.6\%$) across three
seeds, with $\tNC = 237 \pm 6$. For ResNet-20/Fashion-MNIST (image inputs
resized to $32\times 32$ to match the existing ResNet-20 cell), all three
seeds collapse at $\tNC = 660$ with $\featnorm = 1.678 \pm 0.012$
(CV~$=0.71\%$).
The within-pair CV again sits below the $8\%$ bound observed for every
other pair, supporting the claim that the threshold rule generalises beyond
MNIST/CIFAR-10.
For Fashion-MNIST, the Phase-2 trajectories continued to descend past
NC1~$=0.05$ (unlike CIFAR-10/MLP-5, which plateaus): all three seeds reached
NC1~$=0.012$ by the final logged epoch with $\featnorm = 1.169 \pm 0.065$
(CV~$=5.6\%$), still trending downward (Fig.~\ref{fig:fmnist_baseline}).
Taken together, the three MLP-5 datasets place \fnthresh\ in the range
$[0.90, 1.37]$ at the strictest threshold each pair reached, while the
architecture change on MNIST (MLP-5~$\to$~ResNet-20,
Fig.~\ref{fig:summary_panel_d}) moves \fnthresh\ by $458\%$. The architecture
effect is therefore large but strongly dataset-dependent, so \fnthresh\ is best described as
\emph{jointly} determined by architecture and dataset rather than primarily by
either alone, consistent with the non-additive interaction reported above.

\paragraph{Optimiser de-confound.}
Because the main grid pairs Adam with MLP-5 and SGD with ResNet-20, the
architecture effect on \fnthresh\ reported above is, in the main grid,
entangled with optimiser. To separate the two factors we re-ran both
architectures on MNIST under a \emph{matched} Adam optimiser (three seeds each;
Table~\ref{tab:deconfound}). Under matched Adam, MLP-5 collapses at
\fnthresh~$=1.095 \pm 0.039$ while ResNet-20 collapses at
\fnthresh~$=7.468 \pm 0.129$, an architecture effect of $+582\%$, in the same
direction as, and larger than, the $+458\%$ measured in the
(optimiser-confounded) main grid. Holding architecture fixed instead, switching
ResNet-20 from SGD (\fnthresh~$=5.867$, main grid) to Adam raises \fnthresh\ to
$7.468$, a $+27\%$ optimiser effect; this is consistent with optimiser choice
influencing collapse dynamics~\citep{zhao2026optimizer} but is far smaller than
the architecture effect. Running MLP-5 under the ResNet SGD configuration
(learning rate~$0.1$) instead produced degenerate collapse
(\featnorm~$\to 0$, a non-ETF solution), which we exclude. We therefore
conclude that the large architecture effect on \fnthresh\ is not an optimiser
artefact.

\begin{table}[t]
\centering
\caption{Optimiser de-confound on MNIST (three seeds each, strict NC1~$<0.01$;
values mean~$\pm$~std).}
\label{tab:deconfound}
\small
\begin{tabular}{lll}
\toprule
Architecture & Optimiser & \fnthresh \\
\midrule
MLP-5     & Adam (matched)  & $1.095 \pm 0.039$ (reproduces main grid) \\
ResNet-20 & Adam (matched)  & $7.468 \pm 0.129$ \\
ResNet-20 & SGD (main grid) & $5.867$ \\
MLP-5     & SGD (lr $=0.1$) & $\to 0$ (degenerate, non-ETF; excluded) \\
\bottomrule
\end{tabular}
\end{table}

\subsection{H5, Width: Negligible Threshold Shift, Monotone Speed Acceleration}
\label{sec:width}

\begin{figure}[htbp]
  \centering
  \includegraphics[width=\linewidth]{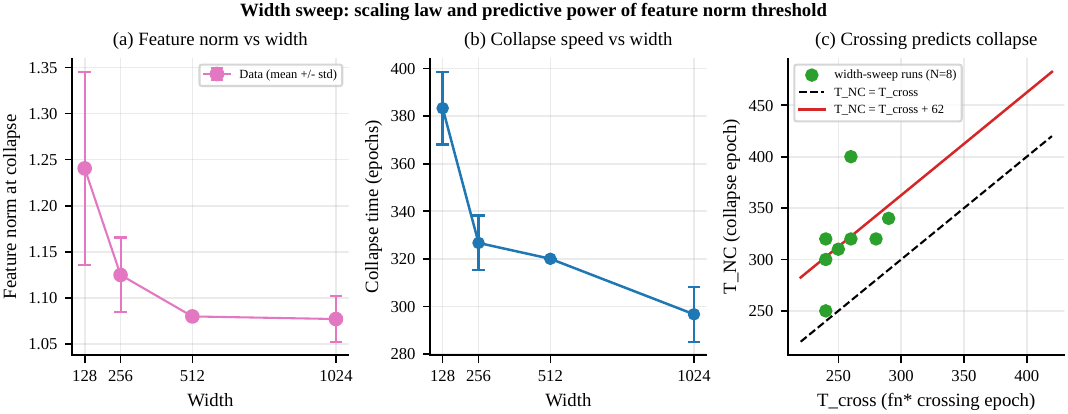}
  \caption{Width sweep (MLP-5, ReLU, $\lambda=10^{-4}$, MNIST):
  (a)~\featnorm\ at $\tNC$, (b)~$\tNC$ vs.\ width, (c)~fn crossing precedes $\tNC$.}
  \label{fig:width}
\end{figure}

Table~\ref{tab:width} and Fig.~\ref{fig:width} report MLP-5 results at
widths $\{128, 256, 512, 1024\}$ (ReLU, $\lambda=10^{-4}$, MNIST, 3~seeds). Width~512
repeats the baseline configuration with independent seeds; its \fnthresh\ agrees
with the depth-5 cell of Table~\ref{tab:depth} within seed variation.

\begin{table}[t]
\centering
\caption{Width sweep (MLP-5, ReLU, $\lambda=10^{-4}$, MNIST, NC1~$<0.01$, $N=3$).}
\label{tab:width}
\begin{tabular}{ccccc}
\toprule
\textbf{Width} & \textbf{Params} & $T_{\mathrm{NC}}$
  & \textbf{\featnorm\ at $\tNC$} & \textbf{CV} \\
\midrule
128  & 0.15M & $383 \pm 15$  & $1.241 \pm 0.105$ & 8.4\% \\
256  & 0.53M & $327 \pm 12$  & $1.125 \pm 0.041$ & 3.6\% \\
512  & 2.10M & $310 \pm 10$  & $1.096 \pm 0.017$ & 1.6\% \\
1024 & 8.39M & $257 \pm 12$  & $1.080 \pm 0.059$ & 5.4\% \\
\bottomrule
\end{tabular}
\end{table}

The width sweep shows three effects.

\textbf{Width has negligible effect on \fnthresh.}
Across the four widths tested, \featnorm\ at $\tNC$ shows only a weak negative
trend: an $8\times$ width increase (0.15M to 8.39M parameters) shifts \featnorm\
by only 13\%, compared to between-pair gaps of 22.7--458\% (Table~\ref{tab:threshold}). With only $N=4$ width
points we do not fit a formal scaling law; the substantive result is the small
magnitude of this shift.
Width within a fixed architecture cannot account for the between-pair gaps;
the completed grid in Section~\ref{sec:threshold} shows those gaps are driven by an architecture
$\times$ dataset interaction.

\textbf{Width monotonically accelerates collapse speed.}
$\tNC$ falls from 383 to 257~epochs across the $8\times$ width range, a 33\%
reduction that is monotone across all four widths.
This monotone effect contrasts with the non-monotone depth effect and suggests
that more channels provide more redundant pathways for \featnorm\ to decay toward
\fnthresh.

\textbf{fn crossing \fnthresh\ precedes $\tNC$ with a mean lead of 62~epochs (width sweep).}
We define $T_{\text{cross}}$ as the first epoch in Phase~2 at which \featnorm\ drops
below \fnthresh\ (the per-width mean collapse value).
In all 8~confirmed width-sweep runs, $T_{\text{cross}} < \tNC$, confirming
the temporal ordering.
The gap $\tNC - T_{\text{cross}}$ has mean $62 \pm 37$~epochs across the 8~runs.
As a descriptive summary of this ordering, $\hat{T}_{\text{NC}} = T_{\text{cross}} + 62$
tracks $\tNC$ with a mean absolute error of 24~epochs (range 2--78~epochs) on these
8~cases; we do not propose it as a practical forecasting rule, since \featnorm\
co-evolves with NC1 rather than leading it (Section~\ref{sec:implications}).
The temporal ordering is not specific to the width sweep: recomputed from the
released per-epoch logs across all 12 strict-threshold (NC1~$<0.01$)
MLP-5/MNIST ReLU runs spanning the width, depth, and weight-decay sweeps
(each evaluated against its own per-condition \fnthresh), $T_{\text{cross}} <
\tNC$ holds in every run (12/12), with mean lead $75 \pm 51$~epochs. The wider
spread relative to the width-only subset reflects cross-condition heterogeneity;
we therefore present \featnorm\ crossing as a consistent qualitative \emph{ordering}
relative to collapse rather than a precise or practically leading predictor; a
head-to-head signal comparison (Section~\ref{sec:implications}) shows the lead is
short and \featnorm\ co-evolves with NC1.

\subsection{Intervention Experiment: \fnthresh\ Is an Attractor, Not a Cause}
\label{sec:intervention}

Among our results, this intervention provides the most direct evidence about the
\emph{causal} status of \fnthresh, and it pre-empts the most natural objection to any
threshold rule. Because we manipulate \featnorm\ directly rather than observing it
passively, the conclusion rests on an intervention rather than on correlation: a
threshold that reliably precedes collapse could in principle be an incidental
by-product of it, and only a direct perturbation can separate the two readings.

To distinguish whether \featnorm\ \emph{causes} collapse or is a \emph{correlated symptom}
of an underlying change in loss geometry, we run an intervention experiment.
At the end of Phase~1, the checkpoint is saved and three Phase-2 conditions are
launched from identical starting weights:
(a)~\emph{control} ($\alpha = 1.0$, no modification);
(b)~\emph{scale-down} ($\alpha = 0.3$): the final hidden layer weights are
multiplied by $\alpha$, reducing \featnorm\ from ${\approx}16$ to
${\approx}4.9$, closer to but still above the collapse value \fnthresh;
(c)~\emph{scale-up} ($\alpha = 3.0$): same weights multiplied by $\alpha$,
pushing \featnorm\ to ${\approx}49$.
All 9 runs (3 conditions $\times$ 3 seeds) confirm collapse within 400 Phase-2
epochs; results are summarised in Table~\ref{tab:intervention}.

\begin{table}[t]
\centering
\caption{Intervention experiment (MLP-5, ReLU, $\lambda=10^{-4}$, MNIST, $N=3$).}
\label{tab:intervention}
\setlength{\tabcolsep}{4.5pt}
\begin{tabular}{lcccc}
\toprule
\textbf{Condition} & $\alpha$ & \textbf{fn after rescaling}
  & $T_{\mathrm{NC}}$ & \textbf{\fnthresh\ at collapse} \\
\midrule
Control    & 1.0 & $16.2$  & $287 \pm 35$ & $1.103 \pm 0.020$ \\
Scale-down & 0.3 & $4.9$   & $317 \pm 6$  & $1.093 \pm 0.033$ \\
Scale-up   & 3.0 & $48.7$  & $317 \pm 46$ & $1.054 \pm 0.049$ \\
\bottomrule
\end{tabular}
\end{table}

The trajectories reveal the mechanism directly. From all three starting points,
the scale-down value (${\approx}4.9$), the control (${\approx}16$), and the
scale-up value (${\approx}49$), the feature norm relaxes to the same \fnthresh\
($\approx 1.06$--$1.10$). The scale-down arm overshoots \emph{below} \fnthresh\ (to
${\approx}0.97$--$1.00$) and is then pulled back up to it, so \fnthresh\ attracts
trajectories from below as well as from above. Collapse-time differences across
arms are only ${\approx}30$~epochs, not statistically significant ($p > 0.2$).

\textbf{Conclusion: \fnthresh\ is an attractor of the MSE + weight-decay gradient flow,
not a causal trigger.}
Collapse does not occur \emph{because} \featnorm\ reaches \fnthresh; rather, both \featnorm\ and NC1
are jointly driven toward their characteristic values by the same underlying
loss geometry.
\featnorm\ remains a reliable temporal marker because both quantities are governed
by the same dynamics: \featnorm\ converges slightly before NC1 collapses, but the
intervention rules out a direct causal chain.
This is consistent with Hypothesis H-M4 (Section~\ref{sec:mechanisms}): \fnthresh\ is the fixed point
of the gradient flow, and the network self-corrects to it from any initialisation.

\subsection{Cross-Architecture Check: a Vision Transformer}
\label{sec:vit}

To test whether the tight, pair-specific feature norm extends beyond the two
architectures of the main grid, we trained a small Vision Transformer (ViT;
embedding dimension 128, depth 6, 4 heads) on MNIST under the same two-phase
CE$\to$MSE protocol and the same NC criteria (3 seeds, AdamW with a short
warmup). The ViT reaches neural collapse at the relaxed criterion: NC1 plateaus
at $0.015$ to $0.021$ and does not cross $0.01$ within 800 epochs, first reaching
NC1~$<0.05$ at epochs 400 to 420, with test accuracy $98.8\%$ and
NC2~$\approx 0.05$. Its classifier-input feature norm concentrates tightly across
seeds, $\fnthresh = 12.67 \pm 0.13$ (CV~$=1.0\%$), comparable to the within-pair
concentration of the ResNet-20 cells.

Two caveats qualify this result. First, the ViT classifier input sits after a
final LayerNorm, which pins its norm near $\sqrt{d}\approx 11.3$, so the value is
not comparable in magnitude to the unnormalised MLP-5 ($1.05$) or ResNet-20
($5.87$) features; the cross-architecture claim rests only on the within-pair
concentration, not the magnitude. Second, and more important, this concentration
is largely imposed by the LayerNorm: the \emph{pre}-LayerNorm CLS norm is not
concentrated across seeds ($907 \pm 141$, CV~$=15.5\%$). The ViT therefore
corroborates that neural collapse and a stable classifier-input feature norm
extend to a transformer, but it does not provide an independent test of an
\emph{emergent} feature-norm threshold of the kind seen in the unnormalised MLP-5
and ResNet-20 models. We report it as a preliminary cross-architecture
observation; full per-epoch logs are released with the code.

\subsection{Protocol Validation: CE Training Alone Is Insufficient}
\label{sec:protocolval}

\begin{figure}[htbp]
  \centering
  \includegraphics[width=\linewidth]{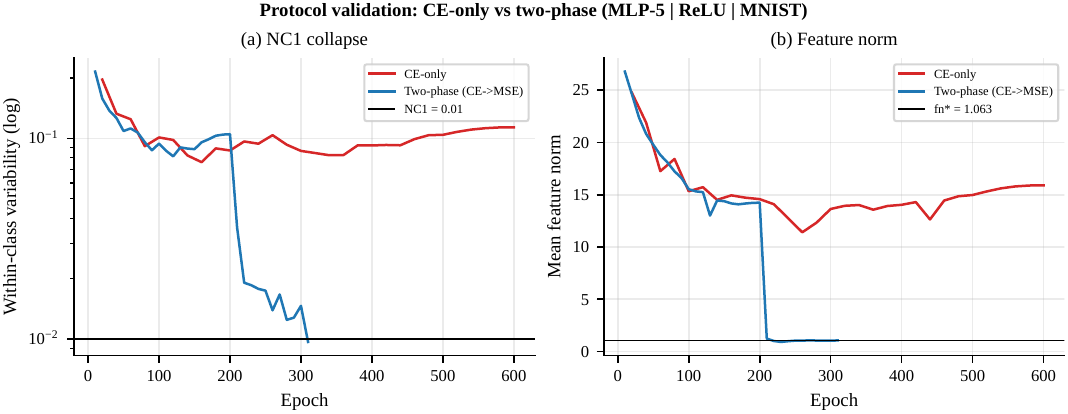}
  \caption{Protocol validation, CE-only vs.\ two-phase training:
  (a)~NC1, (b)~feature norm.}
  \label{fig:ce_vs_twophase}
\end{figure}

Training MLP-5 (ReLU, $\lambda=10^{-4}$, MNIST, seed~0) with CE loss for
600~epochs yields NC1 oscillating between 0.08 and 0.13 throughout
(minimum 0.076, final 0.114).
The feature norm remains between 11 and 25, never approaching the ${\approx}1$
range seen at collapse.
Train accuracy reaches $\geq 99\%$ by epoch~20 and stays there
(Fig.~\ref{fig:ce_vs_twophase}).

\citet{han2021neural} establish NC emergence under CE
asymptotically; \citet{papyan2020prevalence} observe it with
extended training.
In our setup, 600~CE epochs with Adam and cosine annealing are insufficient
to reach this asymptotic regime. The two-phase protocol is therefore
motivated empirically rather than as a free design choice.

\textbf{Implication for the threshold account.}
The CE-only ablation is consistent with, rather than undermining, the central
claim of this paper. The threshold rule predicts that NC onset coincides with
\featnorm\ reaching a $(\text{model},\text{dataset})$-specific equilibrium value
$\fnthresh \approx 1.06$. Under CE alone, \featnorm\ remains an order of
magnitude above this equilibrium (11--25 throughout training), and NC1
correspondingly fails to cross NC1~$<0.01$. The two phenomena are not
independent: collapse below the strict threshold is observed only after the
feature representation approaches the low-norm equilibrium, which in turn is
reached either by (a) an extended training horizon under CE (the asymptotic regime
of \citet{han2021neural}, not feasible in our compute budget),
or (b) an explicit norm-shrinking force such as MSE loss with weight decay.
The two-phase protocol implements option~(b). The CE-only trajectory thus
provides a consistency check: when \featnorm\ stays far from \fnthresh, the
threshold rule predicts no collapse, and that is what we observe.

\section{Discussion}
\label{sec:discussion}

Having established the empirical regularities, we now interpret them: the
structural parallel with grokking (Section~\ref{sec:grokking}), what sets the
threshold value (Sections~\ref{sec:whatdet}--\ref{sec:mechanisms}), the practical
implications (Section~\ref{sec:implications}), and the regimes where the rule
breaks down (Section~\ref{sec:failuremodes}).

\subsection{The Five-Way Parallel with Grokking}
\label{sec:grokking}

The results reveal a structural parallel between NC dynamics and
grokking~\citep{manir2026systematic} that holds across several independent axes.
Both phenomena exhibit:
\begin{enumerate}[leftmargin=*,noitemsep]
\item \textbf{Norm concentration:} \featnorm\ at $\tNC$ is approximately constant
  within each (model, dataset) pair (within-pair CV~$<8\%$).
\item \textbf{Model-specificity:} between-pair gap 22.7--458\% (Table~\ref{tab:threshold}).
\item \textbf{Weight decay controls rate, not threshold:}
  CV~$= 5.9\%$ for \featnorm\ at $\tNC$ across a $10\times$ range of $\lambda$.
\item \textbf{Goldilocks zone:} excessive regularisation prevents the transition.
\item \textbf{Non-monotonic depth effect:} intermediate depth is fastest.
\end{enumerate}

One difference: in grokking, activation changes the rate but not the threshold;
for NC, activation shifts both.
The shift suggests that activations shape the implicit feature geometry of the collapsed
state, determining what \featnorm\ value corresponds to the ETF becoming an attractor, but
we treat this as a hypothesis rather than an established result.

In the sense of \citet{nanda2023progress}, both \featnorm\
(for NC) and the RMS weight norm (for grokking) satisfy the conditions
for a progress measure: they are monotonically associated with the
transition, model-specific, and invariant to optimisation conditions.

Concurrent work by \citet{sakamoto2025explaining} links NC and grokking in the
opposite direction: they use within-class variance (NC1) contraction to derive a
generalisation bound that explains grokking, establishing NC as a sufficient
condition for grokking-style late generalisation, whereas our threshold rule
gives a grokking-style progress measure that precedes NC onset. The two are
complementary: NC1 contraction $\Rightarrow$ grokking generalisation, and
\featnorm\ crossing $\Rightarrow$ NC collapse. Together they
suggest NC and grokking may share a norm-based progress measure marking
the transition, though establishing a common mechanism needs further theory.

\subsection{What Determines the Threshold Value?}
\label{sec:whatdet}

One interpretive observation deserves emphasis. The ResNet-20/MNIST cell
($\fnthresh = 5.867$) is by far the most extreme cell and also collapses the
fastest of any condition ($\tNC = 110$~epochs, versus $660$ for Fashion-MNIST and
CIFAR-10): ResNet-20's ${\sim}270$k parameters are heavily over-specified for
10-class MNIST, reaching an equilibrium geometry, set by batch normalisation (BatchNorm)~\citep{ioffe2015batch}, skip
connections, and feature dimension, that differs fundamentally from the MLP on the
same data. This is consistent with the much larger architecture effect on MNIST
than on Fashion-MNIST or CIFAR-10 (Table~\ref{tab:threshold}). Within-architecture
width shifts \fnthresh\ by at most $13\%$ (Section~\ref{sec:width}), well below the
between-pair gaps, so width is not the explanation; architecture and dataset must
be characterised jointly.

\subsection{A Simplified Unconstrained-Features Account of \fnthresh}
\label{sec:ufm}

We offer a deliberately simplified calculation that rationalises two of our
observations, the order of magnitude of \fnthresh\ and its insensitivity to
small weight decay, within the unconstrained-features model
(UFM)~\citep{zhu2021geometric,zhou2022optimization,tirer2022extended}. It is a
heuristic account intended only to explain the \emph{scale} of \fnthresh\ and its
weak weight-decay dependence, not to derive the measured values; by construction
it does not model the architecture-dataset interaction of Section~\ref{sec:threshold}.

Consider $K$ balanced classes at exact collapse, so each class is represented by
a single penultimate feature $\mathbf{h}_c$ and classifier row $\mathbf{w}_c$.
Take the symmetric simplex-ETF ansatz $\mathbf{w}_c=\alpha\,\mathbf{u}_c$ and
$\mathbf{h}_c=\beta\,\mathbf{u}_c$, where $\{\mathbf{u}_c\}$ is a simplex ETF
($\mathbf{u}_c^\top\mathbf{u}_c=1$, $\mathbf{u}_j^\top\mathbf{u}_c=-\tfrac{1}{K-1}$
for $j\neq c$). The logits for a class-$c$ example are
$\mathbf{w}_j^\top\mathbf{h}_c=\alpha\beta$ for $j=c$ and $-\alpha\beta/(K-1)$
otherwise. With one-hot MSE targets and $L_2$ regularisation $\lambda$ on the
weights and features, the per-class objective is
\begin{equation}
J=(\alpha\beta-1)^2+\frac{\alpha^2\beta^2}{K-1}
   +\frac{\lambda}{2}\bigl(\alpha^2+\beta^2\bigr).
\end{equation}
By symmetry $\alpha=\beta=:r$, and stationarity $\mathrm{d}J/\mathrm{d}r=0$
(for $r\neq0$) gives
\begin{equation}
r^2=\frac{K-1}{K}\Bigl(1-\frac{\lambda}{2}\Bigr),
\qquad
\fnthresh=\lVert\mathbf{h}_c\rVert=r
       =\sqrt{\tfrac{K-1}{K}\bigl(1-\tfrac{\lambda}{2}\bigr)} .
\end{equation}
Two consequences follow. First, for $K=10$ this predicts
$\fnthresh\approx\sqrt{0.9}\approx0.95$ in units of the one-hot target scale, the
same order as the measured MLP-5/MNIST value ($1.05$). Second, and more
informatively, \fnthresh\ depends on $\lambda$ only through the factor
$(1-\lambda/2)$: across our weight-decay range ($10^{-5}$ to $10^{-4}$) the
predicted change is below $0.005\%$, and even a $1000\times$ increase to
$10^{-2}$ moves it by only $0.25\%$. The empirical near-invariance of \fnthresh\
to weight decay (Section~\ref{sec:threshold}) is therefore expected in the small-$\lambda$
regime: the feature scale is set primarily by the target magnitude, with weight
decay a sub-dominant correction. This calculation contains no architecture or
dataset variable and hence says nothing about the between-pair gaps; the
architecture-dataset interaction remains an empirical finding beyond the UFM
abstraction.

\subsection{Candidate Mechanisms (Hypotheses)}
\label{sec:mechanisms}

Our experiments establish \emph{what} happens to \fnthresh\ under varying conditions but
not \emph{why}. We outline four candidate mechanisms, each making a testable
prediction. These are intended as hypothesis generators rather than explanatory
claims, and none is established by the present experiments.

\textbf{H-M1: Positive homogeneity correlates with the equilibrium \fnthresh\ magnitude.}
Positively homogeneous activations (ReLU, LeakyReLU) admit a scale symmetry that
MSE-plus-weight-decay breaks, which may select a lower feature scale. As detailed
in Section~\ref{sec:activation}, the homogeneous activations cluster near
\featnorm~$\approx 1.0$ while smooth non-homogeneous GELU and SiLU sit far higher;
but this ordering does not survive at a common NC1 criterion, so H-M1 is
unsupported by the present evidence and retained only as a hypothesis, to be tested
by a fixed-threshold ReLU control $\mathrm{ReLU}(x-\tau)$ ($\tau>0$) that removes
homogeneity while preserving piecewise-linearity.

\textbf{H-M2: Competing depth effects, representational capacity versus accumulated implicit regularisation, produce the non-monotonic collapse speed.}
Each hidden layer adds to the effective weight-decay regularisation on the
features, so deeper networks may slow the rate at which \featnorm\ decays toward
\fnthresh\ without changing \fnthresh\ itself.
This would explain the non-monotonic depth effect: depth-3 is fast enough to
benefit from multi-layer representational capacity but not so deep as to incur
severe implicit regularisation; depth-7 is slowed but eventually reaches the same
equilibrium \fnthresh.
\emph{Testable prediction}: holding the total weight-decay budget constant (i.e.,
$\lambda/\text{depth}$ fixed) should partially collapse the non-monotonic depth
effect on $\tNC$ while leaving \fnthresh\ unchanged.

\textbf{H-M3: BatchNorm~\citep{ioffe2015batch} and skip connections~\citep{he2016deep} raise the equilibrium \fnthresh\ in ResNets.}
ResNet-20's much higher \fnthresh\ (5.867 vs.\ 1.052 for MLP-5 on MNIST) suggests
architectural constraints beyond activation shape the equilibrium. Complementing
the over-parameterisation interpretation in Section~\ref{sec:whatdet}, BatchNorm
(which normalises intermediate activations) and skip connections (which maintain
feature scale) are the structural features through which the extra capacity could
sustain a higher feature norm at collapse while preserving the ETF structure.
\emph{Testable prediction}: removing BatchNorm from ResNet-20 while retaining skip
connections (or vice versa) should shift \fnthresh\ toward the MLP-5 value, identifying
which structural element is the primary contributor.

\textbf{H-M4: \fnthresh\ is the fixed point of the MSE + weight-decay gradient flow},
set by the balance between the MSE alignment force and the weight-decay shrinkage
force. The intervention experiment (Section~\ref{sec:intervention}) supports this
directly. \emph{Open question}: whether the fixed point is derivable in closed
form from the loss landscape and feature-map Jacobian, yielding \fnthresh\ as a
function of width and depth.

\subsection{Implications and Future Directions}
\label{sec:implications}

The results point to three concrete directions, each directly grounded in the
empirical findings.

\textbf{1.~Feature norm as a stable landmark of collapse.}
Because \featnorm\ concentrates tightly at \fnthresh\ and collapse coincides with
\featnorm\ reaching it, \fnthresh\ provides a cheap, stable summary of the
collapsed state: one forward pass yields \featnorm, versus the full NC1/NC2/NC3
computation. We caution against reading this as a practical early-warning signal.
A head-to-head comparison of candidate signals (feature norm, training loss,
training error, weight norm, gradient norm) on MLP-5/MNIST, restricted to Phase~2
to remove a loss-scale artefact at the CE$\to$MSE boundary, shows that none
provides a meaningful lead over NC1: all cross their onset-calibrated thresholds
within a few tens of epochs of collapse, and \featnorm\ does not lead the others.
This aligns with the intervention result (Section~\ref{sec:intervention}):
because \fnthresh\ is an attractor and \featnorm\ and NC1 are driven by the same
gradient flow, \featnorm\ crossing is a descriptive temporal ordering, not an
independent precursor, and the lead is short and varies across conditions
($75\pm51$~epochs over 12 runs, Section~\ref{sec:width}). Its value is therefore
as an inexpensive proxy for the collapsed-state geometry, useful where NC is
desirable (transfer learning~\citep{galanti2021role,li2024transfer}, few-shot and
class-incremental recognition~\citep{yang2023neural}), rather than as a signal
that anticipates collapse earlier than training loss or accuracy would.

\textbf{2.~Principled control of collapse timing.}
Weight decay, width, and depth are three predictable levers that adjust \tNC\ while
leaving \fnthresh\ largely invariant: weight decay shifts \tNC\ by up to 90~epochs
over a $10\times$ range in a three-regime structure (too low slows, optimal fastest,
too high prevents collapse); width accelerates collapse monotonically by up to 33\%
over an $8\times$ parameter range; and depth is non-monotonic, with intermediate
values fastest. Because these levers modulate the \emph{rate} of approach rather
than \fnthresh\ itself, practitioners can plan training duration from the early
\featnorm\ trajectory and the known \fnthresh.

\textbf{3.~A norm-dynamics perspective on representational theory.}
Where prior NC theory characterises the equilibrium geometry, our results suggest
the \emph{trajectory toward} it is characterised by \featnorm\ approaching \fnthresh, a
far simpler object, reframing the question from ``what is the geometry of the NC
attractor?'' to ``what sets the \featnorm\ at which it becomes reachable, and why
does this depend jointly on architecture and dataset?'' The candidate mechanisms
below and the grokking parallel suggest norm dynamics may be a unifying language
for delayed representational reorganisation.

\subsection{When the Threshold Rule Breaks Down}
\label{sec:failuremodes}

The threshold rule is not universal, and its boundaries are as informative as its
successes. We collect here the regimes in which collapse, or the rule itself,
fails within our experiments.

\textbf{Excessive weight decay prevents collapse.} At $\lambda \geq 5\times10^{-4}$
(MLP-5, ReLU, MNIST), no seed reaches the collapse criterion: NC1 stalls at
${\approx}0.03$ and \featnorm\ never settles at \fnthresh\
(Table~\ref{tab:wd}). It marks the upper boundary of the Goldilocks zone
(Section~\ref{sec:wd}): the rule holds \emph{within} the collapsing regime, but
sufficient regularisation suppresses collapse entirely.

\textbf{High class counts.} Extending ResNet-20 to CIFAR-100 (100 classes) did
not produce a valid collapse under any of three optimiser configurations; the
CE$\to$MSE transition destabilised the Phase-1 representation, which we attribute
to a mismatch between the two-phase protocol and 100-class sparse one-hot MSE
targets. We therefore cannot report a CIFAR-100 \fnthresh, and the rule remains
untested above ten classes.

\textbf{Smooth non-homogeneous activations near the criterion.} SiLU does not
cleanly cross a fixed NC1 threshold within the Phase-2 budget (its NC1 plateaus
just above $0.05$), so \fnthresh\ must be estimated at the epoch of minimum
Phase-2 NC1 rather than read off a crossing (Section~\ref{sec:activation},
Table~\ref{tab:act}). The fixed-threshold definition that works for ReLU does not
transfer unchanged to plateauing activations.

\textbf{The weakest cell.} MLP-5/CIFAR-10 reaches only the relaxed NC1~$<0.05$
criterion, with a wide confidence interval (CV~$=7.3\%$, 95\% CI width $36\%$ of
the mean); we label it preliminary and rely on it for no central claim
(Section~\ref{sec:threshold}).

\textbf{Prediction is descriptive, not a causal lever.} The feature-norm lead is a
temporal ordering whose magnitude varies across conditions
($75\pm51$~epochs over 12 runs), and the intervention experiment
(Section~\ref{sec:intervention}) shows that rescaling \featnorm\ does not reliably
shift the collapse epoch: \featnorm\ self-corrects to \fnthresh\ and $\tNC$
differences are not significant ($p>0.2$). The threshold is therefore best read as
a stable landmark, not a control knob for collapse timing.

\subsection{Limitations}
\label{sec:limitations}

Three of the five activations tested (Tanh, GELU, and SiLU) use a relaxed
NC1 collapse criterion compared to ReLU and LeakyReLU's NC1~$<0.01$.
For Tanh and GELU we use NC1~$<0.05$; this is not a free parameter but a
necessary adjustment, as both saturate above 0.01 within practical horizons,
and a stricter threshold would simply censor otherwise valid runs.
SiLU does not cleanly cross any fixed NC1 threshold within 1000 Phase-2
epochs (min NC1 across seeds: $0.032$--$0.059$); we therefore report
\featnorm\ at the epoch of minimum Phase-2 NC1, a stable estimator of the
equilibrium under plateau dynamics (Section~\ref{sec:activation}, Table~\ref{tab:act}).
The SiLU definition differs methodologically from the other activations and
should be interpreted accordingly; the qualitative conclusion that SiLU's
\featnorm\ lies in the GELU regime (well above the ReLU/LeakyReLU regime) is
robust to the choice of estimator (mean fn over the plateau, fn at end of
Phase~2, and fn at minimum NC1 all agree within $5\%$, see Section~\ref{sec:activation}).
Because all comparisons are made within-activation when drawing conclusions,
these threshold differences do not affect any qualitative claims.

Each ResNet-20 configuration uses $N=3$ seeds. The within-cell variance is
negligible (CV~$\leq 0.6\%$, 95\% $t$-intervals within $3\%$ of the mean), over an
order of magnitude smaller than the reported effect sizes ($+458\%$, $+68\%$), so
the small sample does not affect any ranking or conclusion.

Two MLP-5 cells require the relaxed criterion: MLP-5/CIFAR-10 ($N=3$,
CV~$=7.3\%$, 95\%~CI $[0.738,1.063]$, width 36\% of the mean) and
MLP-5/Fashion-MNIST ($N=3$, CV~$=5.6\%$, 95\%~CI $[1.176,1.559]$, width
28\% of the mean). Neither network reaches NC1~$<0.01$ within the Phase-2
budget; both are therefore reported at NC1~$<0.05$. We explicitly label the
MLP-5/CIFAR-10 cell as preliminary and do not rely on it for any central
claim; removing it entirely leaves all conclusions unchanged.
For Fashion-MNIST, the trajectory continues to descend well past NC1~$=0.05$
(Section~\ref{sec:threshold}), which supports rather than weakens the threshold
claim.

For ResNet-20 on both CIFAR-10 and Fashion-MNIST, all seeds meet the collapse
criterion only at the final logged epoch (660), so the reported $\tNC = 660$ is a
logging-horizon bound and true collapse may occur slightly earlier.
This induces only a bounded, one-sided timing uncertainty and does not affect any
comparative statements.

The width sweep uses only $N=4$ width values, too few to fit a reliable scaling
law, so we report the 13\% feature-norm variation and the monotone collapse
speed-up as descriptive trends only, and interpret the 13\% as an upper bound on
the within-architecture width effect rather than a precise estimate.

The architecture-dataset grid is evaluated on six (model, dataset) cells
(MLP-5 on MNIST, Fashion-MNIST, and CIFAR-10; ResNet-20 on MNIST,
Fashion-MNIST, and CIFAR-10).
These six cells are sufficient to show a non-additive interaction in the settings evaluated, though not to
fully parameterise it; additional cells would allow the interaction surface to be
characterised more completely. Given the magnitude of the observed effects, this
broader mapping is unlikely to overturn the qualitative finding of non-additivity.

All experiments use a two-phase training protocol. Alternative protocols may
shift absolute values, but overturning our conclusions would require changes
comparable to the between-architecture gaps, which prior evidence does not
suggest.

Optimiser is tied to architecture in the main grid (Adam for MLP-5, SGD for
ResNet-20), so the architecture effect is confounded with optimiser there; the
matched-optimiser de-confound (Section~\ref{sec:threshold},
Table~\ref{tab:deconfound}) resolves this, with the architecture effect persisting
under common Adam and the residual optimiser effect an order of magnitude smaller.
Because optimiser choice itself influences NC dynamics~\citep{zhao2026optimizer},
our findings apply within a fixed optimiser per architecture; calibrating
\fnthresh\ across optimiser families (e.g.\ AdamW with decoupled weight decay)
remains future work.

High-class-count settings remain untested: our CIFAR-100 attempt did not collapse
under the present protocol (Section~\ref{sec:failuremodes}), and we leave these
settings to future work.

Our claims are bounded by the datasets and architectures studied: three
$\leq$10-class image benchmarks (MNIST, Fashion-MNIST, CIFAR-10) on MLP-5 and
ResNet-20. Within this scope the threshold rule is consistent across depths,
weight decays, widths, activations, and seeds; Fashion-MNIST, added in revision,
shows the concentration rule holding outside the original MNIST/CIFAR-10 setting
(within-pair CV~$=5.6\%$). We make no claim beyond these regimes: the specific
\fnthresh\ values and the rule itself remain to be tested at higher class counts
(our CIFAR-100 attempt did not collapse under the present protocol), on
large-scale or higher-resolution data such as ImageNet, on non-vision modalities,
and on architectures such as Transformers, a natural direction for future work.
The effects we did observe give no indication of a qualitative reversal, though
confirming this remains an empirical question.
Finally, neural collapse is primarily a property of the training set, and its
connection to test-set performance is debated~\citep{hui2022limitations};
accordingly, our threshold characterises collapse \emph{timing}, a
training-dynamics phenomenon, and we make no claim about generalisation.

\section{Conclusion}
\label{sec:conclusion}

This paper provides a controlled characterisation of neural collapse \emph{onset},
a regime prior work has largely left unexamined in favour of the collapsed
equilibrium. Across five experiments we establish that onset is marked by one
measurable quantity: the mean penultimate feature norm at collapse concentrates
tightly within each (model, dataset) pair (within-pair CV~$<8\%$ across depths,
weight decays, widths, and seeds, for every pair tested across MNIST,
Fashion-MNIST, and CIFAR-10) and marks the onset of collapse. This concentration
is a reproducible structural property of the dynamics, not incidental noise
reduction.

Completing the (architecture)$\times$(dataset) grid reveals a non-additive
interaction (Table~\ref{tab:threshold}, Section~\ref{sec:threshold}): the
architecture effect on \fnthresh\ depends on which dataset is held fixed,
and the dataset effect depends on which architecture is held fixed, with
conditional effect sizes differing by up to a factor of ${\approx}20$.
A matched-optimiser de-confound (Table~\ref{tab:deconfound}) confirms this
interaction is architectural rather than an optimiser artefact, and the
under-prediction of the largest cell by any multiplicative model means \fnthresh\
for a new (model, dataset) pair must be measured directly, not extrapolated from
marginals.

Four further regularities constrain the phenomenon: depth affects collapse speed
non-monotonically; activation jointly sets both collapse speed and \fnthresh;
width accelerates collapse monotonically; and weight decay controls only the rate
of approach to \fnthresh, except beyond a sharp cutoff ($\lambda \geq
5\times10^{-4}$) where collapse is prevented entirely.

An intervention experiment (Section~\ref{sec:intervention}) further shows that, when penultimate
features are rescaled over a tenfold range, the feature norm returns
to the same \fnthresh. This identifies \fnthresh\ as an
attractor of the gradient flow rather than a direct causal trigger, consistent
with its role as a stable landmark of the collapsed state.

Taken together, these regularities form a coherent structural signature that
closely parallels the weight-norm threshold phenomenon observed in
grokking~\citep{manir2026systematic} (see Section~\ref{sec:grokking} for the detailed
five-way comparison and one acknowledged exception).
The correspondence is sufficiently specific to motivate a unified theoretical
investigation of neural collapse and grokking: both settings exhibit
threshold-controlled transitions, delayed reorganisation, and sharp regime
boundaries, suggesting that norm-threshold dynamics may be a general
mechanism underlying delayed representational phase changes in neural
networks.

\paragraph{Outlook.}
A predictive theory of collapse onset remains open. It should derive \fnthresh\
from architecture and data rather than measuring it per pair, explain why the
architecture--dataset dependence is non-additive, and place the feature-norm
attractor and the grokking weight-norm threshold within a single account of
delayed representational transitions. Our results delimit what such a theory must
reproduce.

\paragraph{Practical takeaway.}
The threshold gives a cheap, stable summary of the collapsed state. Rather than
computing the full NC1/NC2/NC3 metrics, which require forward passes over the
entire training set, a practitioner can read the mean feature norm \featnorm\ (a
single forward pass): collapse coincides with \featnorm\ contracting to the
pair-specific value \fnthresh, the only quantity that must be calibrated per
(model,~dataset) pair. Because \featnorm\ co-evolves with NC1 and does not lead
training loss or accuracy (Section~\ref{sec:implications}), it is a proxy for the
collapsed-state \emph{geometry} rather than an early-warning signal: an inexpensive,
single-pass quantity a practitioner can track in place of the full NC1/NC2/NC3
metrics.

\section*{Data availability}
The code and per-run data are publicly available at
\url{https://github.com/Rupawheatly/NC}; a single notebook recomputes every reported
quantity and regenerates every figure.

\section*{CRediT authorship contribution statement}
\textbf{Anamika Paul Rupa:} Conceptualization, Methodology, Software, Formal
analysis, Investigation, Data curation, Writing -- original draft, Writing --
review \& editing, Visualization.

\section*{Declaration of competing interest}
The author declares that she has no known competing financial interests or personal
relationships that could have appeared to influence the work reported in this paper.

\section*{Funding}
This research did not receive any specific grant from funding agencies in the
public, commercial, or not-for-profit sectors.

{\small
\setlength{\bibsep}{1pt plus 0.3ex}
\bibliographystyle{elsarticle-num-names}
\bibliography{nc_references}
}

\end{document}